\documentclass{article}

\usepackage{arxiv}

\usepackage[utf8]{inputenc} % allow utf-8 input
\usepackage[T1]{fontenc}    % use 8-bit T1 fonts
\usepackage{hyperref}       % hyperlinks
\usepackage{url}            % simple URL typesetting
\usepackage{booktabs}       % professional-quality tables
\usepackage{amsfonts}       % blackboard math symbols
\usepackage{nicefrac}       % compact symbols for 1/2, etc.
\usepackage{microtype}      % microtypography
\usepackage{lipsum}
\usepackage{graphicx}
\usepackage{algorithm}
\usepackage{algorithmic}
\usepackage{array}
\usepackage{enumitem}
\usepackage{multirow}
\usepackage{amsmath}
\usepackage{subcaption}
\usepackage{hyperref}
\def\bng{\bngx}

\font\bngx=bang10

\def\*#1*#2{o\null{#2}{#1}}

\def\sh#1{\setbox0=\hbox{#1}%
     \kern-.02em\copy0\kern-\wd0
     \kern.04em\copy0\kern-\wd0
     \kern-.02em\raise.0433em\box0 }
\graphicspath{ {./images/} }

\title{ANCHOLIK-NER: A Benchmark Dataset for Bangla Regional Named Entity Recognition}

\author{
 Bidyarthi Paul \\
  Department of Computer Science and Engineering\\
  Southeast University\\
  Dhaka, Bangladesh\\
  \texttt{bidyarthipaul01@gmail.com}\\
  \And
 Faika Fairuj Preotee\\
  Department of Computer Science and Engineering\\
  Southeast University\\
  Dhaka, Bangladesh\\
  \texttt{faikafairuj2001@gmail.com}\\
  \And
 Shuvashis Sarker\\
  Department of Computer Science and Engineering\\
  Southeast University\\
  Dhaka, Bangladesh\\
  \texttt{shuvashisofficial@gmail.com}\\
  \And
 Shamim Rahim Refat \\
  Department of Computer Science and Engineering\\
  Ahsanullah University of Science and Technology\\
  Dhaka, Bangladesh\\
  \texttt{n.a.refat2000@gmail.com}\\
  \And
 Shifat Islam\\
  Department of Computer Science and Engineering\\
  Bangladesh University of Engineering and Technology\\
  Dhaka, Bangladesh\\
  \texttt{shifat.islam.buet@gmail.com}\\
  \And
 Tashreef Muhammad \\
  Department of Computer Science and Engineering\\
  Southeast University\\
  Dhaka, Bangladesh\\
  \texttt{tashreef.muhammad@seu.edu.bd}\\
  \And
 Mohammad Ashraful Hoque \\
  Department of Computer Science and Engineering\\
  Southeast University\\
  Dhaka, Bangladesh\\
  \texttt{ashraful@seu.edu.bd}\\
  \And
 Shahriar Manzoor \\
  Department of Computer Science and Engineering\\
  Southeast University\\
  Dhaka, Bangladesh\\
  \texttt{smanzoor@seu.edu.bd}\\
}

\begin{document}
\maketitle

\begin{abstract}
Named Entity Recognition (NER) in regional dialects is a critical yet underexplored area in Natural Language Processing (NLP), especially for low-resource languages like Bangla. While NER systems for Standard Bangla have made progress, no existing resources or models specifically address the challenge of regional dialects such as Barishal, Chittagong, Mymensingh, Noakhali, and Sylhet, which exhibit unique linguistic features that existing models fail to handle effectively. To fill this gap, we introduce ANCHOLIK-NER, the first benchmark dataset for NER in Bangla regional dialects, comprising 17,405 sentences distributed across five regions. The dataset was sourced from publicly available resources and supplemented with manual translations, ensuring alignment of named entities across dialects. We evaluate three transformer-based models—Bangla BERT, Bangla BERT Base, and BERT Base Multilingual Cased—on this dataset. Our findings demonstrate that BERT Base Multilingual Cased performs best in recognizing named entities across regions, with significant performance observed in Mymensingh with an F1-score of 82.611\%. Despite strong overall performance, challenges remain in region like Chittagong, where the models show lower precision and recall. Since no previous NER systems for Bangla regional dialects exist, our work represents a foundational step in addressing this gap. Future work will focus on improving model performance in underperforming regions and expanding the dataset to include more dialects, enhancing the development of dialect-aware NER systems.

% ANCHOLIK-NER is a linguistically diverse dataset for Named Entity Recognition (NER) in Bangla regional dialects, capturing variations across Sylhet, Chittagong, Barishal, Noakhali, and Mymensingh. The dataset has around 17,405 sentences, 3,481 sentences per region. The data was collected from two publicly available datasets and through web scraping from various online newspapers, articles. To ensure high-quality annotations, the BIO tagging scheme was employed, and professional annotators with expertise in regional dialects carried out the labeling process. The dataset is structured into separate subsets for each region and is available in CSV format. Each entry contains textual data along with identified named entities and their corresponding annotations. Named entities are categorized into ten distinct classes: Person, Location, Organization, Food, Animal, Colour, Role, Relation, Object, and Miscellaneous. This dataset serves as a valuable resource for developing and evaluating NER models for Bangla dialectal variations, contributing to regional language processing and low-resource NLP applications. It can be utilized to enhance NER systems in Bangla dialects, improve regional language understanding, and support applications in machine translation, information retrieval, and conversational AI.
\end{abstract}

% keywords can be removed

\section{Introduction}
Named Entity Recognition (NER) is a part of Information Extraction (IE) that focuses on identifying and classifying named entities in unstructured text – names of persons, organizations, locations, dates, numbers and other specific terms. By extracting entities, Named Entity Recognition (NER) transforms raw text into structured data, enabling tasks like information retrieval, question answering, and knowledge graph construction. NER was first introduced in the Sixth Message Understanding Conference (MUC-6) which emphasized extracting structured information from free-form natural language \cite{grishman1996message}. Standard entity types like Person, Organization and Location were defined in MUC-6 and later refined in MUC-7 with clearer annotation guidelines and evaluation metrics \cite{chinchor1998appendix}. Multilingual adaptation efforts like The Conventions d'annotations en Entit{\'e}s Nomm{\'e}es-ESTER project addressed language specific annotation conventions \cite{le2004conventions}. Today NER is a foundation step in NLP pipelines supporting applications like machine translation, automatic summarization, topic detection and recommendation systems \cite{nadeau2007survey}. So NER is essential for systems that want to extract semantic meaning and support decision making in both academic and industrial NLP tasks.

\begin{figure}[htbp]
    \centering
    \includegraphics[width=0.8\textwidth]{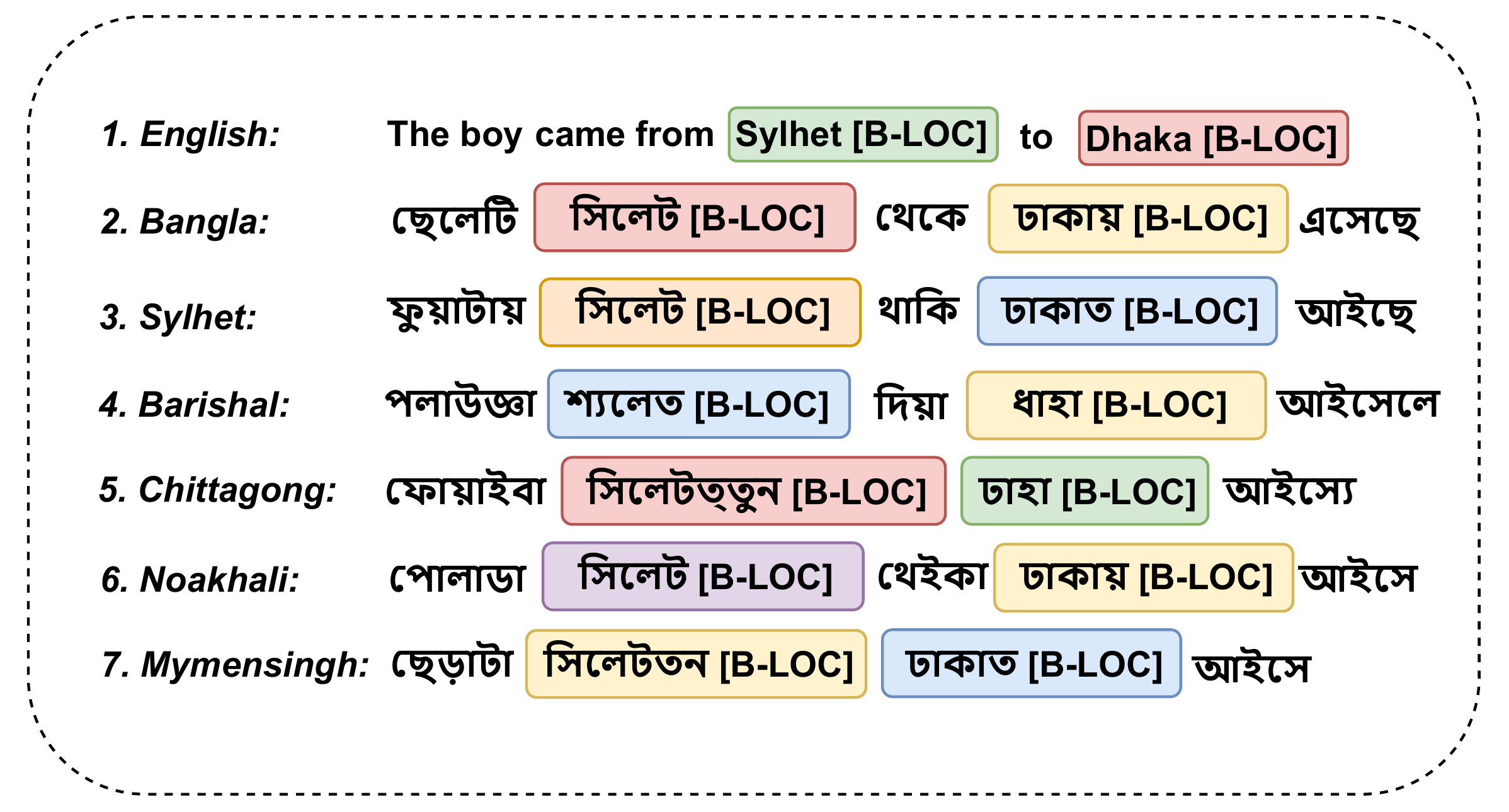}
    \caption{Regional NER examples along with Standard Bangla and English}
    \label{fig:intro}
\end{figure}

Initial approaches to the NER task were rule-based, relying on manually crafted lexicons, grammars and pattern-matching heuristics to extract entities from domain-specific corpora \cite{mikheev1999knowledge, mikheev1999named, palmer1997statistical}. These systems were limited in scalability, language coverage and generalizability as they were dependent on domain knowledge and rigid linguistic rules. As the limitations of handcrafted systems became apparent—especially in adapting to different languages and sentence structures—researchers moved towards data-driven approaches. This led to the adoption of statistical and machine learning models that required annotated datasets but offered much better adaptability and inference capabilities. Among the earliest machine learning models used for NER were Hidden Markov Models (HMMs), which modelled sequences of labels with probabilistic transitions and emissions \cite{brants2000tnt, de2003memory}. These were soon followed by more powerful models like Conditional Random Fields (CRFs) \cite{lafferty2001conditional, li2003rapid} which could capture both contextual features and label dependencies across sequences. In parallel, Maximum Entropy (ME) \cite{ratnaparkhi1996maximum} and Maximum Entropy Markov Models (MEMMs) \cite{mccallum2000maximum} were explored for their flexibility in incorporating different features. Support Vector Machines (SVMs) \cite{kudo2001chunking, yamada2003statistical} were introduced as robust classifiers for NER, especially when combined with rich feature representations and kernel methods. To further improve performance, Ensemble Modeling techniques \cite{florian2003named, alam2011named} were proposed, combining multiple learners to mitigate overfitting and improve generalization. With the increasing availability of annotated NER datasets across multiple languages, multilingual and cross-lingual models started to emerge \cite{abdallah2012integrating} allowing knowledge transfer between related language families. The evolution of NER reached a new milestone with the rise of deep learning, especially recurrent neural networks (RNNs) and their variants like Long Short-Term Memory (LSTM) networks. These models could capture long-range dependencies in text and when combined with CRFs produced state-of-the-art results in sequence labelling tasks \cite{huang2015bidirectional}. The introduction of Bi-directional LSTM-CRF architectures further improved entity recognition by leveraging both past and future context \cite{lample2016neural}. The most significant breakthrough came with the development of pre-trained transformer-based models like BERT and multilingual BERT (mBERT) \cite{devlin2019bert}  which enabled deep contextualized word embeddings and fine-tuning for downstream tasks including NER. Variants like RoBERTa \cite{liu2019roberta}, and XLM-R \cite{conneau2019unsupervised} for low-resource and performance across languages. In recent years, NER esearch has continued to evolve rapidly. GLiNER \cite{zaratiana2023gliner}, a lightweight generative NER framework, outperformed ChatGPT in zero-shot evals and used way less memory. ToNER \cite{jiang2024toner}, introduced type-oriented generation and showed great results in prompt-based NER. KoGNER \cite{zhang2025kogner} added knowledge graph distillation to transformer-based models and improved entity classification in biomedical NER. FewTopNER \cite{bouabdallaoui2025fewtopner} proposed a few-shot, topic-aware method and improved cross-lingual generalization for low-resource NER. For noisy and informal data like social media, SEMFF-NER \cite{li2024method} used multi-scale feature fusion and syntactic information to handle irregular sentence structures better. Large Language Model Cooperative Communication (LLMCC) \cite{yang2024intelligent} tried cooperative learning between large language models for NER and got better extraction consistency through model coordination. Named Entity Recognition for All (NER4All) \cite{hiltmann2025ner4all} addressed challenges in historical and multilingual corpora and introduced methods for adaptive entity recognition across genre and domain shifts. Multimodal NER \cite{xu2025enhancing} also made progress with Adaptive Mixup methods that fused image-text representations for better recognition in visually grounded contexts. These recent developments show NER is still going strong through interdisciplinarity and scalability.

In recent years, a lot of progress has been made in Bangla Named Entity Recognition (NER) systems which is one of the most spoken yet under-resourced language in the world. Initial works used statistical models like Conditional Random Fields (CRFs) and Support Vector Machines (SVMs) with handcrafted features on small scale annotated corpora \cite{ekbalnamed}. Later Bi-LSTM and Bi-LSTM-CRF were introduced which offered better contextual modeling and yielded better results \cite{karim2019step}. With the rise of pre-trained language models, transformer based approaches like multilingual BERT (mBERT) and BanglaBERT were used for Bangla NER and further improved the results by fine-tuning on domain specific tasks \cite{bhattacharjee2021banglabert}. Along with modeling techniques, several tools and resources were developed including BNLP toolkit \cite{leon2022complexity}, BnNER and BNERCorpus \cite{haque2023b}. Recently B-NER released one of the largest Bangla NER dataset to date along with benchmark evaluations \cite{haque2023b}. Gazetteer-Enhanced improved the recognition performance by using K-Means infused CRF with BanglaBERT embeddings \cite{farhan2024gazetteer}. TriNER \cite{dhamaskar2025triner} explored multilingual NER by training models across Hindi, Bengali and Marathi and enabled cross-lingual learning for low resource languages. Despite all the progress in Named Entity Recognition (NER) for Bangla, there are still many limitations that hinder the development of high performing and generalizable models. Most of the existing Bangla NER models are trained on small or synthetic datasets that don’t capture the linguistic richness and real world variability of the language. As a result, these models perform poorly on diverse contexts like informal speech, domain specific corpora and most critically regional dialects. One of the most overlooked yet impactful challenge in Bangla NER is the lack of support for regional dialects which are widely spoken in Bangladesh in areas like Chittagong, Sylhet, Barishal, Khulna and Mymensingh. These dialects vary greatly from standard Bangla in terms of vocabulary, grammar, pronunciation and semantics making it difficult for models trained only on formal Bangla corpora to identify and classify named entities in dialectal text. This gap is particularly problematic for practical applications like social media analysis, regional news summarization, public health communication and localized digital services – domains where dialectal usage is frequent and essential. Moreover, the absence of large scale, gold standard annotated corpora along with inconsistencies and mismatches in translation based datasets further restricts model performance. Transformer based models like BanglaBERT and mBERT, although promising, still lag behind those for high resource languages due to limited morphological, cultural and linguistic adaptation. The exclusion of dialectal diversity not only leads to poor recognition and inaccurate information extraction but also introduces a form of linguistic bias that undermines inclusivity. Given the socio-cultural importance of regional speech in Bangladesh, development of dialect aware NER models is not just a technical improvement – it’s a linguistic and ethical imperative. Addressing these challenges requires a systematic effort to build large, high quality and regionally diverse datasets along with adaptable and context sensitive modeling approaches that can serve all segments of Bangla speaking populations.

To get beyond these limitations our approach takes a data centric view and emphasizes linguistic inclusivity and technical precision. Rather than focusing on model architecture or fine tuning strategies we focus on enhancing the quality, diversity and dialectal relevance of the training data used in Bangla NER. While previous works have focused on standard Bangla we explore underrepresented regional linguistic contexts by curating dialect aware datasets and translation corpora that preserve named entity integrity. We investigate how dialectal variation affects NER and how it can be modeled through better data design. Figure~\ref{fig:intro} shows us an example of Dialect variations in Bangla language in NER. We also address the challenges in translation based NER where entity misalignments and inconsistencies degrade the model performance by incorporating structured correction and validation strategies. Our approach is based on empirical benchmarking using pre-trained transformer models so that each proposed resource is practical and measurable in terms of downstream effectiveness. Here are the key contributions of our work:

\begin{itemize}
    \item \textbf{Regional Bangla Dialect-Aware NER Dataset} \\
    We present the first NER dataset for Bangla that includes annotations from five major regional dialects—\textit{Chittagong, Sylhet, Barishal, Noakhali}, and \textit{Mymensingh}—addressing the lack of dialectal coverage in existing resources.

    \item \textbf{Hybrid Translation Dataset with Entity Alignment} \\
    We construct a parallel corpus between standard Bangla and regional dialects, ensuring that \textit{named entities} are consistently preserved and aligned across translations.
    
    \item \textbf{Anomaly Detection and Data Refinement} \\
    We apply systematic anomaly detection and correction to remove noisy or inconsistent data, ensuring high-quality hybrid datasets for robust model training.

    \item \textbf{Evaluation with Transformer Models} \\
    We benchmark our datasets using transformer models like \textbf{Bangla BERT}, \textbf{Bangla BERT Base}, and \textbf{BERT Base Multilingual Cased}, demonstrating improved performance in dialect-aware NER tasks. The Bangla BERT model achieved the highest F1-score of 0.82413 in Mymensingh at epoch 20, while Bangla BERT Base performed well with an F1-score of 0.80916 at epoch 10. The BERT Base Multilingual Cased model also showed strong performance in the same region, achieving an F1-score of 0.82611 at epoch 20.
\end{itemize}

By addressing these critical challenges in Bangla NER including dialectal diversity, limited annotated resources, and translation inconsistencies—our approach ensures better generalization, higher entity recognition accuracy, and improved adaptability across regional variations. This leads to a more robust and scalable NER framework suitable for diverse Bangla language applications. This paper is structured as follows: Section \ref{Problem_Description} describes the problem in detail, highlighting the core challenges addressed in this study. Section \ref{Background} provides a comprehensive literature review of existing research and methods in Bangla NER. Section \ref{Dataset_Preparation} discusses dataset preparation, including \ref{Data_Collection} Data Collection and \ref{Data_pre-processing_and_Tokenization} data pre-processing and tokenization strategies used to curate and refine the datasets. Section \ref{Data_Availability} explains the availability and characteristics of the data resources developed. Section \ref{Methodology} outlines the proposed methodology, detailing the dataset construction, anomaly detection techniques, and benchmarking approaches. Section \ref{Result_Analysis} presents the result analysis and performance evaluation of the proposed framework. Finally, Section \ref{Conclusion_and_Future_work} concludes the paper by summarizing the key findings, discussing limitations, and outlining directions for future research.

\section{Problem Description}
\label{Problem_Description}

Let \( n \) represent the number of words in a sentence. The sentence can be denoted as a set \( s = \{ w_1, w_2, w_3, \dots, w_n \} \), where each word is treated as an individual token. The output is a corresponding set of tags \( t = \{ t_1, t_2, t_3, \dots, t_n \} \), where each tag \( t_i \) belongs to the set of predefined entity tags:
\[
t_i \in \{ \text{B-PER}, \text{I-PER}, \text{B-LOC}, \text{I-LOC}, \text{B-ORG}, \text{I-ORG}, \text{B-REL}, \text{I-REL}, \text{B-FOOD}, \text{I-FOOD}, \text{B-ANI}, \text{I-ANI}, O \}.
\]
The context of the sentence must be considered when assigning tags to each token.

\section{Related Works}
\label{Background}

With the rapid expansion of user-generated content, social networking platforms, and digital services, Named Entity Recognition (NER) has become a critical task in the field of Natural Language Processing (NLP), serving as a foundational step in information extraction pipelines. NER facilitates structured understanding of text by identifying key entities such as people, places, organizations, and temporal expressions—playing a vital role in applications like news aggregation, knowledge graph construction, recommendation systems, and biomedical research. In high-resource languages such as English, Spanish, and German, the combination of large annotated corpora, linguistically rich resources, and transformer-based models like BERT and RoBERTa has enabled researchers to achieve state-of-the-art F1 scores nearing 95\%. %{https://nlpprogress.com/english/named_entity_recognition.html}
However, this success has not translated evenly across languages. Low-resource languages, including Bangla, suffer from a scarcity of labeled datasets, inconsistent annotation schemes, and limited availability of pretrained models tailored to linguistic nuances. Moreover, existing Bangla NER systems often focus on coarse-grained entities (like person or location), neglecting fine-grained and domain-specific categories. Recent efforts have started addressing these gaps by developing domain-adaptable NER systems. 

Recent advancements in Named Entity Recognition (NER) for low-resource languages and domain-specific applications reveal a shared focus on large-scale dataset creation, domain-adaptive pretraining, and leveraging large language models (LLMs). In the context of Chinese NER, Yao et al. \cite{yao2024agcner} introduced AgCNER, a substantial annotated dataset for agricultural diseases and pests, alongside AgBERT, a domain-specific model achieving an impressive 94.34\% F1-score. Complementing this, researchers \cite{jia2020entity} constructed Chinese NER datasets from Internet novels and financial reports, adopting a semi-supervised, entity-enhanced BERT pretraining approach that integrated lexicon-level knowledge into deep contextual embeddings. Both studies demonstrate the critical role of large annotated datasets and domain-specific adaptations for enhancing NER across diverse Chinese text genres.
A similar pattern emerges in Indonesian NER research, where Yulianti et al. \cite{yulianti2024named} and Khairunnisa et al. \cite{khairunnisa2024improving} explored distinct but complementary approaches. The authors \cite{yulianti2024named} developed IndoLER, a legal-domain NER dataset with 1,000 annotated court documents and 20 fine-grained entity types, demonstrating that transformer-based models like XLM-RoBERTa and IndoRoBERTa outperformed traditional architectures(BiLSTM-CRF), achieving F1-scores up to 0.929. In contrast, Khairunnisa et al. \cite{khairunnisa2024improving} introduced IDCrossNER, a cross-domain Indonesian dataset derived via semi-automated translation from English, and applied GPT-based augmentation and transfer learning, significantly improving NER performance in limited-data scenarios. These studies highlight the role of multilingual resources and domain adaptation in addressing Indonesian NER challenges.
In the medical NER domain, spanning across Chinese, Bangla, and English-language social media, the focus shifts toward extracting structured health information from unstructured text. Ge et al. \cite{ge2024reddit} introduced Reddit-Impacts, a novel dataset capturing clinical and social impacts of substance use from Reddit posts, annotated with 30 low-frequency but critical entity types. Their study demonstrated that few-shot learning models like DANN and GPT-3.5 outperformed traditional methods such as BERT and RoBERTa, highlighting the challenges of sparse entity recognition in health discourse. Aligning with this, Muntakim et al. \cite{muntakim2023banglamedner} developed BanglaMedNER, the largest Bangla medical NER corpus with over 117,000 tokens across three key categories (Chemicals \& Drugs, Diseases \& Symptoms). Their BiLSTM-CRF model achieved a 75\% macro F1-score, demonstrating the feasibility of applying deep learning techniques in low-resource, specialized medical contexts. %Notably, Yao et al.’s (2024) AgCNER dataset, while focused on agricultural diseases in Chinese, complements these efforts by showcasing how domain-specific pretraining and large-scale annotations enhance NER performance in specialized health-related fields.

In multilingual settings, cross-lingual transfer has proven effective, as seen in NaijaNER \cite{oyewusi2021naijaner} , which covered five Nigerian languages and demonstrated that multilingual models outperform language-specific counterparts. This trend extends to Indo-Aryan languages, where annotated datasets for Assamese \cite{pathak2022asner} and Purvanchal languages (Bhojpuri, Maithili, Magahi) \cite{mundotiya2023development} aligned entity labels with Hindi NER corpora, facilitating knowledge transfer among related languages. Such alignment not only enriches low-resource datasets but also enhances model adaptability across linguistically similar regions. Similar strategies have been applied to West Slavic languages like Upper Sorbian and Kashubian, leveraging Czech and Polish corpora for cross-lingual learning \cite{torge2023named}, and in Uzbek, where annotated datasets with BIOES tagging schemes improved entity boundary detection in legal texts \cite{mengliev2025comprehensive}. In the medical domain, cross-domain adaptation and linguistic diversity remain critical. An annotated corpus of over 117,000 tokens for medical NER (MNER) \cite{muntakim2023banglamedner} exemplifies structured dataset creation, while another study \cite{khan2023nervous}  addressed informal health discourse in Consumer Health Questions (CHQs), capturing dialectal variations and achieving a modest F1-score of 56.13\% with BanglishBERT. Collectively, these works reflect a unified strategy of leveraging cross-lingual alignment, shared annotation schemes, and domain-aware embeddings to advance NER performance in low-resource and domain-specific contexts.

Although substantial progress has been made in NER in terms of dataset quality, annotation consistency, and domain adaptability. A challenge is observed in Urdu NER, where Anam et al. \cite{anam2024deep} highlighted the scarcity of annotated resources and addressed out-of-vocabulary (OOV) issues by leveraging FastText and Floret embeddings with RNN variants. Their approach successfully captured sub-word patterns, yet the dependence on benchmark datasets restricts model adaptability across broader domains.
In specialized areas like biomedical NER, the problem of limited labeled data becomes even more pronounced. Gao et al. \cite{gao2021pre} tackled this by adopting a transfer learning and self-training framework, pretraining models on extensive biomedical corpora like SemMed and MedMentions. While their approach reduced the dependency on manual annotations, it also highlighted the reliance on domain-specific corpora for maintaining high performance. Extending further, Yan et al. \cite{yan2025chinese} addressed the complexity of Chinese medical NER by integrating RoBERTa-wwm-ext, BiLSTM, multi-head attention, and a gated context-aware mechanism (GCA) to effectively model long-range dependencies and entity relationships. However, their evaluation on datasets like MCSCSet and CMeEE also pointed out challenges in achieving consistent performance across different medical subdomains.

A few research works have significantly contributed to the development of Bangla NER datasets, though challenges related to entity diversity, annotation quality, and model adaptability persist.  For instance, Lima et al. \cite{lima2023novel} developed a large-scale Bangla NER dataset containing over 1 million tokens across six entity types; however, despite its size, the dataset struggled with class imbalance, particularly dominated by non-entity tokens. To mitigate this, the authors integrated character-level embeddings within a hybrid CNN-BiLSTM/GRU-CRF model, demonstrating improved recognition of complex entity patterns in morphologically rich Bangla text. Karim et al. \cite{karim2019step} initiated this line of research by creating a dataset of over 71,000 sentences, annotated across four core entity types—person, location, organization, and object—using the IOB tagging scheme. Their model, which integrated Densely Connected Networks (DCN) with BiLSTM for sequence labeling, achieved an F1-score of 63.37\%. Building on the need for improved datasets and architectures, Rifat et al. \cite{rifat2019bengali} expanded the scope by annotating 96,697 tokens and benchmarking several deep learning models, including BLSTM+CNN and BGRU+CNN. Their work marked a shift away from traditional machine learning models (e.g., HMM, CRF, SVM) towards neural architectures that leveraged character-level embeddings. Despite this, their best-performing model (BGRU+CNN) attained an F1-score of 72.66\%, though annotation inconsistencies and dataset imbalance remained key limitations. 
%indicating persistent limitations tied to annotation inconsistencies and imbalanced datasets, much of which stemmed from semi-automated labeling processes.

%Karim et al. initiated this line of research by creating a Bangla NER dataset with over 71,000 sentences annotated across four entity types, using a DCN-BiLSTM model that achieved an F1-score of 63.37\%. However, the limited entity scope and moderate performance highlighted the challenges of Bangla’s morphological complexity. Expanding this work, Rahman Rifat et al. annotated 96,697 tokens and evaluated deep learning models like BLSTM+CNN and BGRU+CNN, shifting from traditional machine learning to neural architectures. Their best model achieved a 72.66\% F1-score, though annotation inconsistencies and dataset imbalance remained key limitations.
Addressing these shortcomings, Haque et al. \cite{haque2023b} introduced B-NER, which not only increased annotation reliability but also broadened entity diversity to eight categories, including complex entities like events and artifacts. Unlike previous works, their dataset of 22,144 sentences was fully manually annotated, achieving a Kappa score of 0.82, ensuring high inter-annotator agreement. This rigorous annotation framework, combined with evaluations using IndicbnBERT, yielded a Macro-F1 score of 86\%, significantly outperforming earlier models and datasets.

\section{Dataset Preparation}
\label{Dataset_Preparation}

Natural Language Processing (NLP) has emerged as a foundational area within artificial intelligence, aiming to equip machines with the ability to understand, interpret, and generate human language. Among the various tasks in NLP, Named Entity Recognition (NER) remains particularly challenging. As a sequence labeling problem, NER involves detecting and classifying named entities—such as persons, locations, organizations, and more—within unstructured text. The availability of large, high-quality annotated datasets has significantly advanced NER systems in high-resource languages.

In the context of Bangla, a language spoken by millions yet historically considered low-resource, there has been a noticeable shift in recent years. Efforts to improve Bangla NLP have led to the development of benchmark datasets for several key tasks, including abstractive text summarization\cite{summarization}, question answering\cite{QA}, authorship classification\cite{authorship}, and machine translation\cite{machine}. These initiatives mark a promising step toward reducing the resource gap. However, NER-specific resources in Bangla remain limited, and existing datasets are primarily focused on Standard Bangla, neglecting the diverse range of regional dialects spoken across the country. Bangladesh is home to a range of regional dialects, including Sylhet, Chittagong, Barishal, Noakhali, and Mymensingh, each with its own distinct phonological, lexical, and syntactic characteristics. These dialects, while widely spoken, are significantly underrepresented in computational linguistics. The lack of annotated resources for these varieties presents a major obstacle for building inclusive and dialect-aware NLP systems. In the context of NER, this gap is particularly pronounced, as dialectal variations can alter named entity boundaries, spellings, and even categories, reducing the accuracy of models trained solely on Standard Bangla.

\begin{figure}[htbp]
    \centering
    \includegraphics[width=\linewidth]{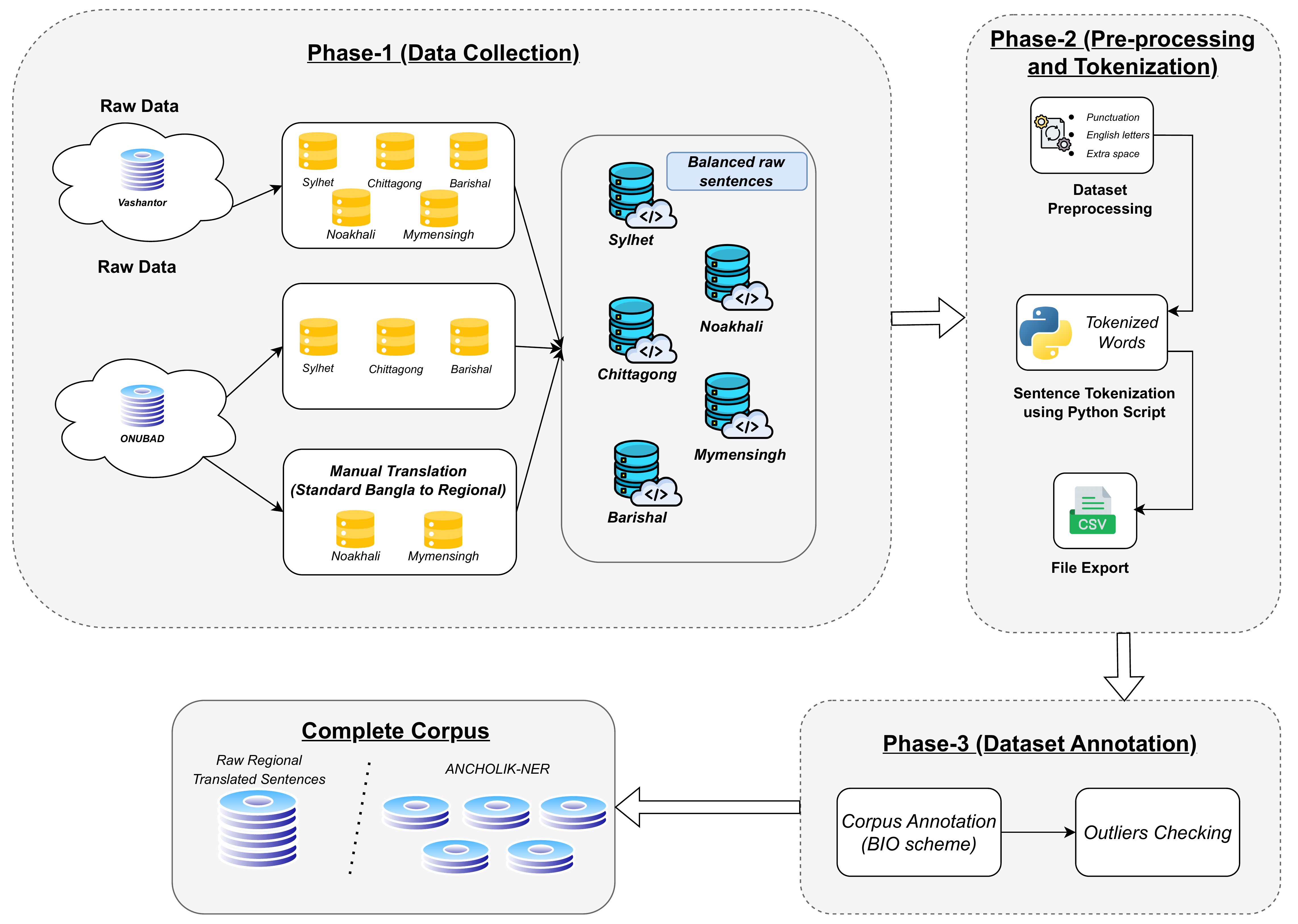}
    \caption{Development of ANCHOLIK-NER: A Systematic Pipeline for Dataset Creation}
    \label{fig:data_dev}
\end{figure}

To address this gap, our work introduces ANCHOLIK-NER, the first benchmark dataset for Named Entity Recognition in Bangla regional dialects. This dataset is designed to support the development of NER models capable of understanding dialect-specific named entities across Sylhet, Chittagong, Barishal, Noakhali, and Mymensingh. The dataset construction process follows a systematic pipeline, ensuring high annotation quality through expert linguists familiar with these dialects. Figure~\ref{fig:data_dev}, illustrates the pipeline for the development of our dataset.

\subsection{Data Collection}
\label{Data_Collection}

Preparation of the ANCHOLIK-NER dataset starts with it's data collection phase, which was compiled from a combination of publicly available sources and manual collection efforts. A total of 17,405 sentences were gathered across five Bangla regional dialects: Sylhet, Chittagong, Barishal, Noakhali, and Mymensingh. The majority of the data—12,500 sentences—was sourced from the Vashantor corpus \cite{faria2023vashantor}. To construct a unified and consistent dataset, we merged the training, validation, and test splits from the original corpus into a single collection, while 2,940 sentences were extracted from the ONUBAD dataset\cite{sultana2025onubad}.To enhance dialectal diversity and ensure balanced representation, an additional 1,965 sentences were manually obtained through regional dialect translation. As the ONUBAD dataset contains data for only three regions—Sylhet, Barishal, and Chittagong—we first collected the corresponding Standard Bangla versions of the sentences. These were then manually translated into the Noakhali and Mymensingh dialects to complete the dataset. Table\ref{tab:sentence_count}, summarizes the distribution of the data sources.

\begin{table}[htbp]
    \centering
    \caption{Distribution of Sentences Across Different Data Sources for Bangla Regional Dialects in the ANCHOLIK-NER Dataset}
    \renewcommand{\arraystretch}{1.1}
    \begin{tabular}{p{4cm}  r}
        \toprule
        \textbf{Sources} & \textbf{Sentence Count} \\
        \midrule
        Vashantor\cite{faria2023vashantor} & 12,500 \\
        ONUBAD\cite{sultana2025onubad} &  2940 \\
        Manual Translation & 1965 \\ 
        \bottomrule
        \textbf{Total} & 17,405 \\
        \bottomrule
    \end{tabular}
    
    \label{tab:sentence_count}
\end{table}

\subsection{Data pre-processing and Tokenization}
\label{Data_pre-processing_and_Tokenization}

As the dataset was constructed by combining content from multiple sources—including existing datasets like Vashantor\cite{faria2023vashantor} and ONUBAD\cite{sultana2025onubad}, as well as manually translated sentences—many noisy and irrelevant components were present in the raw text. These included inconsistencies such as punctuation anomalies, English numerals, and other non-Bangla elements that needed to be cleaned to ensure quality for downstream tasks like Named Entity Recognition (NER). The manual translation process, particularly for dialectal variations, introduced further irregularities. Since the ONUBAD dataset covers only three regions (Sylhet, Barishal, and Chittagong), we first collected standard Bangla versions of each sentence and then translated them into regional dialects, including Noakhali and Mymensingh through our hired annotators, specialized in regional dialects. These transformations sometimes resulted in mixed-language tokens, requiring a dedicated cleaning step. Some Initial data cleaning was manually done to get rid of the Punctuation anomaly and Mixed-language token, which has been shown in Table~\ref{tab:initial}.

\begin{table}[htbp]
\centering
\caption{Sentence structure conversion by separating punctuation.}
\begin{tabular}{|>{\bfseries}l|p{5cm}|p{5cm}|}
\hline
\textbf{Conversion Type} & \textbf{Before} & \textbf{After} \\
\hline
\multirow{2}{*}{Punctuation Anomaly} 
& {\bng tu{I}  du{I}j/ja  bhat Ha{I}eya en??} 
& {\bng tu{I}  du{I}j/ja  bhat Ha{I}eya en?} \\
& (What did you eat rice with in the afternoon??) 
& (What did you eat rice with in the afternoon?) \\
\hline
\multirow{2}{*}{Mixed Language or Script} 
& {\bng br/tman} govt. {\bng Aenk pRkl/p bas/tbaJn kerech.} 
& {\bng br/tman sorkar Aenk pRkl/p bas/tbaJn kerech.} \\
& (The current government has implemented many projects.) 
& (The current government has implemented many projects.) \\
\hline
\end{tabular}
\label{tab:initial}
\end{table}

To apply further pre-processing, we developed a custom data cleaning and tokenization pipeline using a Python script to make it automated. This pre-processing step involved:

\begin{enumerate}
    \item Extraneous symbols using Python's regular expressions (re) module.
    \item Separating punctuation marks from words to ensure accurate tokenization.
\end{enumerate}

English numerals were replaced with Bengali equivalents, which was done manually by the annotators. The tokenization was designed to prepare the data for annotation in a format suitable for NER tagging scheme. 

\begin{table}[htbp]
    \centering
    \caption{Dataset structure for Sylhet region after pre-processing and tokenization phase (Followed for all 5 regions)}
    \renewcommand{\arraystretch}{1.1}
    \begin{tabular}{p{4cm} p{4cm}}
        \toprule
        \textbf{Sentence \#} & \textbf{Word (Sylhet Region)}\\
        \midrule
        1 & {\bng phuyaTay} \\
        NaN & {\bng iselT} \\
        NaN & {\bng thaik} \\
        NaN & {\bng Dhakat} \\
        NaN & {\bng Aa{I}ech} \\
        \bottomrule
    \end{tabular}
    
    \label{tab:token}
\end{table}

The tokenization process, as described in Algorithm 1, is designed to convert raw Bangla Regional sentences into a structured, tokenized format suitable for downstream annotation and analysis. The algorithm begins by reading the input dataset, which is in CSV format, containing Bangla Regional. 5 different CSVs contains the Raw sentences of 5 different regions. Each sentence is processed one at a time. If the sentence is non-empty, it is passed through a regular expression-based tokenizer to split it into individual tokens. These tokens are typically Bangla Regional words, extracted by filtering out punctuation and whitespace. The regular expression ensures that unwanted characters such as punctuation marks and special symbols are excluded from the tokens. Each sentence is assigned a unique identifier (e.g., "Sentence: 1"), and the first token is associated with this identifier. The remaining tokens from the sentence are listed in subsequent rows, with their sentence identifier left as None (NaN), ensuring compatibility with common sequence labeling formats such as the CoNLL-style layout\cite{sang2003introduction}. Finally, the tokenized output is written to a CSV file with two columns: one for the sentence identifier and the other for the individual words as shown in Table~\ref{tab:token}. 

\begin{algorithm}
\caption{Process Bangla Regional Sentences from Data File}
\begin{algorithmic}[1]
\STATE \textbf{Input:} Read the data file (CSV) containing Bangla Regional sentences
\STATE newRows $\leftarrow$ [ ]
\STATE sentenceID $\leftarrow$ 1
\FOR{each row in the data file}
    \STATE sentence $\leftarrow$ extract sentence from column
    \IF{sentence is not empty}
        \STATE tokens $\leftarrow$ split the sentence using regular expression
        \STATE newRows $\leftarrow$ newRows + [``Sentence: '' + sentenceID, tokens[0]]
        \FOR{$i = 1$ to length(tokens) - 1}
            \STATE newRows $\leftarrow$ newRows + [None, tokens[$i$]]
        \ENDFOR
        \STATE sentenceID $\leftarrow$ sentenceID + 1
    \ENDIF
\ENDFOR
\STATE \textbf{Output:} Write newRows to CSV file with columns: Sentence \#, Word
\end{algorithmic}
\end{algorithm}

\subsection{Annotation Scheme}
Regarding the tagging scheme, various approaches such as BILUO, BIO, BIO2, IO, and BIOES have been proposed for tagging named entities in NER tasks \cite{alshammari2021impact}. The choice of tagging scheme is crucial, as it impacts the granularity and accuracy of the entity identification process. For languages like Bangla, which is a post-positional language, the BIO tagging scheme has been widely adopted by researchers \cite{gonzalez2024named}. Our proposed B-NER dataset follows this convention, with each entity chunk tagged as follows: the first token of an entity is labeled as "B-entity name" (Beginning), and subsequent tokens in the entity are labeled as "I-entity name" (Inside). All other tokens that do not belong to any named entity are tagged as "O-entity name" (Outside). 

\begin{itemize}
    \item B (Beginning): Beginning of a multi-word entity.
    \item I (Inside): Inside a multi-word entity.
    \item O (Outside): Outside any entity.
\end{itemize}

This approach is consistent with the BIO tagging scheme first introduced by Ramshaw and Marcus \cite{ramshaw1999text}, ensuring that our dataset adheres to widely accepted conventions in the field of NER. 
% Table~\ref{tab:tagging}, illustrates a detailed description of this tagging style.

\begin{table}[htbp]
    \centering
    \caption{BIO Tagging Scheme with Examples for Named Entity Recognition in Bangla Regional Dialects}
    \renewcommand{\arraystretch}{1.1}
    \begin{tabular}{|
        >{\centering\arraybackslash}p{2cm}|
        >{\centering\arraybackslash}p{3.5cm}|
        p{8.5cm}|}
        \toprule
        \textbf{Tag} & \textbf{Example} & \textbf{Explanation} \\
        \midrule
        \textit{B-LOC} & \begin{tabular}{c}{\bng lalbag eklLar} \\(Lalbagh Fort)\end{tabular}  & Tags the starting word of a Location name\\ 
        \midrule
        \textit{I-LOC} & \begin{tabular}{c}{\bng lalbag eklLar} \\(Lalbagh Fort)\end{tabular} & Tags the inside of a multi-word Location name \\ 
        \midrule
        \textit{B-PER} & \begin{tabular}{c}{\bng il{O}enl emis} \\(lionel Messi)\end{tabular} & Tags the starting word of a Person name \\ 
        \midrule
        \textit{I-PER} & \begin{tabular}{c}{\bng il{O}enl emis} \\(lionel Messi)\end{tabular} & Tags the inside of a multi-word Person name \\ 
        \midrule
        \textit{B-ORG} & \begin{tabular}{c}{\bng pabilk ibshWibdYaley}\\ (Public University)\end{tabular} & Tags the starting word of an Organization name \\ 
        \midrule
        \textit{I-ORG} & \begin{tabular}{c}{\bng pabilk ibshWibdYaley}\\ (Public University)\end{tabular} & Tags the inside of a multi-word Organization name \\ 
        \midrule
        \textit{B-FOOD} & \begin{tabular}{c}{\bng emaregr maNNGs} \\(Chicken Meat)\end{tabular} & Tags the starting word of a Food name \\ 
        \midrule
        \textit{I-FOOD} & \begin{tabular}{c}{\bng emaregr maNNGs} \\(Chicken Meat)\end{tabular}& Tags the inside of a multi-word Food name \\ 
        \midrule
        \textit{B-ANI} & \begin{tabular}{c}{\bng murigr ba{I}c/ca} \\(Chicken Chick)\end{tabular}& Tags the starting word of an Animal name \\ 
        \midrule
        \textit{I-ANI} & \begin{tabular}{c}{\bng murigr ba{I}c/ca} \\(Chicken Chick)\end{tabular}& Tags the inside of a multi-word Animal name \\ 
        \midrule
        \textit{B-COL} & \begin{tabular}{c}{\bng kala rNNG}\\ (Black Color)\end{tabular}& Tags the starting word of a Color name \\ 
        \midrule
        \textit{I-COL} & \begin{tabular}{c}{\bng kala rNNG} \\(Black Color)\end{tabular}& Tags the inside of a multi-word Color name  \\ 
        \midrule
        \textit{B-ROLE} & \begin{tabular}{c}{\bng kaemr ma{I}ya} \\(Domestic Worker)\end{tabular}& Tags the starting word of a Role title \\ 
        \midrule
        \textit{I-ROLE} & \begin{tabular}{c}{\bng kaemr ma{I}ya} \\(Domestic Worker)\end{tabular} & Tags the inside word of a multi-word Role title \\
        \midrule
        \textit{B-REL} & \begin{tabular}{c}{\bng brh Aaphar}\\ (Elder Sister)\end{tabular} & Tags the starting word of a Relationship label\\ 
        \midrule
        \textit{I-REL} & \begin{tabular}{c}{\bng brh Aaphar}\\ (Elder Sister)\end{tabular} & Tags the inside word of a multi-word Relationship label \\
        \midrule
        \textit{B-OBJ} & \begin{tabular}{c}{\bng lakDir cula}\\ (Wood Stove)\end{tabular} & Tags the starting word of a Physical Object name \\ 
        \midrule
        \textit{I-OBJ} & \begin{tabular}{c}{\bng lakDir cula}\\ (Wood Stove)\end{tabular} & Tags the inside word of a multi-word Physical Object name \\
        \midrule
        \textit{O} & \begin{tabular}{c}{\bng iden, mushikl} \\(Daytime, Difficult)\end{tabular} & Tokens that do not belong to any named entity class \\
        \bottomrule
    \end{tabular}
    
    \label{tab:tagging}
\end{table}

\subsection{Annotators Identity}

A total of ten annotators were recruited for the annotation task, with two annotators assigned to each of the five regional dialects: Chittagong, Sylhet, Barishal, Noakhali, and Mymensingh. The group consisted of both graduate and undergraduate students with academic backgrounds in Linguistics and Natural Language Processing (NLP). Their experience levels varied, ranging from 1 to 4 years, with an average age of 25.7 years. All annotators were native speakers of the respective dialects and were selected to ensure balanced dialectal representation in the corpus. Prior to annotation, each annotator underwent a training session to familiarize themselves with the guidelines and the annotation interface. Detailed information of the annotators is presented in Table~\ref{tab:annotators}.

\begin{table}[htbp]
\centering
\caption{Comprehensive Overview of Annotators’ Background and Expertise}
\begin{tabular}{|c|c|c|c|c|c|}
\hline
\textbf{Annotator} & \textbf{Region} & \textbf{Role} & \textbf{Age} & \textbf{Research Field} & \textbf{Experience} \\ \hline
Annotator 1 & Chittagong & Graduate & 28 & Linguistics & 3 years \\
Annotator 2 & Chittagong & Under-graduate & 22 & NLP & 1 year \\ \hline
Annotator 3 & Sylhet & Graduate & 30 & NLP & 4 years \\
Annotator 4 & Sylhet & Under-graduate & 23 & Linguistics & 2 years \\ \hline
Annotator 5 & Barishal & Under-graduate & 24 & NLP & 1 year \\
Annotator 6 & Barishal & Under-graduate & 25 & Linguistics & 1 year \\ \hline
Annotator 7 & Khulna & Graduate & 27 & NLP & 3 years \\
Annotator 8 & Khulna & Under-graduate & 23 & Linguistics & 2 years \\ \hline
Annotator 9 & Mymensingh & Graduate & 29 & NLP & 4 years \\
Annotator 10 & Mymensingh & Under-graduate & 22 & Linguistics & 1 year \\ \hline
\end{tabular}
\label{tab:annotators}
\end{table}

\subsection{Data Annotation}

For the annotators, we have provided detailed guidelines to ensure consistency and accuracy in tagging named entities. These guidelines help in categorizing and labeling various types of named entities within the text, following a standardized approach. The Location (LOC) tag refers to geographical entities, including cities, towns, landmarks, rivers, mountains, and other notable physical areas. The Person (PER) tag is applied to the names of individuals, covering public figures, common citizens, and fictional characters. The Organization (ORG) tag represents formal entities such as companies, educational institutions, government bodies, and NGOs. Food (FOOD) includes consumable items, ranging from raw ingredients to prepared dishes and beverages. The Animal (ANI) tag covers species and breeds within the biological kingdom Animalia, including both domesticated and wild animals. Color (COL) applies to terms that describe specific shades, hues, or composite colors. The Role (ROLE) tag is used for job titles or professional positions, including general roles like "teacher" or "manager," as well as more specific functional titles. Relationship (REL) encompasses familial and social connections, such as terms like "mother," "father," "friend," and "colleague." The Object (OBJ) tag refers to tangible, physical items, including everyday objects, tools, and machines. Finally, the Non-Entity (O) tag is used for words that do not belong to any named entity class, such as common nouns, verbs, and adjectives that are not associated with specific entities. Table~\ref{tab:tagging}, illustrates the annotation guidelines for our BIO scheme.

\begin{figure}[htbp]
    \centering
    % First Image - Inter-Annotator Agreement
    \begin{minipage}{0.45\textwidth}
        \centering
        \includegraphics[width=\linewidth]{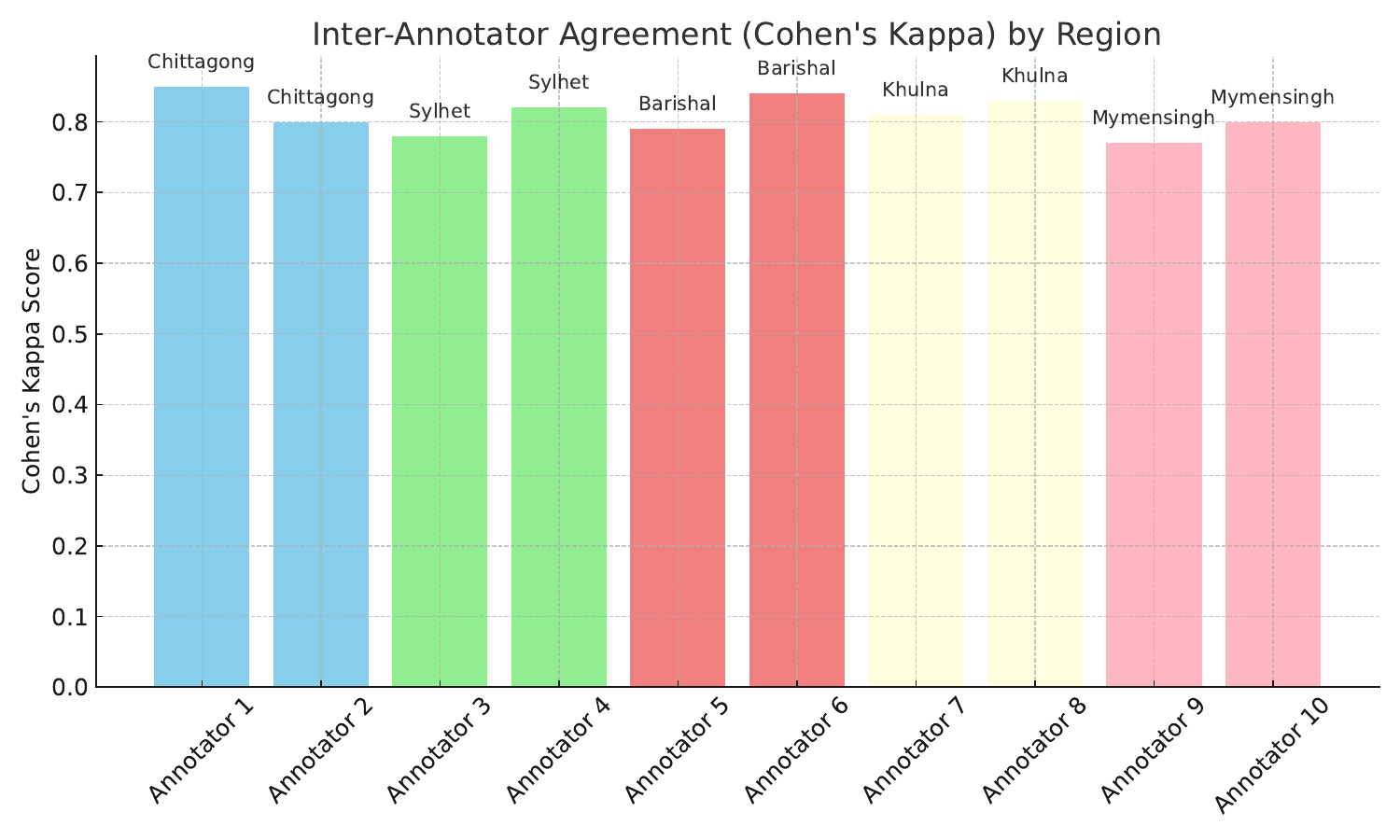} % Replace with actual image file path
        \caption{Inter-Annotator Agreement (Cohen's Kappa) across different regions.}
        \label{fig:kappa}
    \end{minipage}
    \hfill
    % Second Image - Average Tagging Speed
    \begin{minipage}{0.45\textwidth}
        \centering
        \includegraphics[width=\linewidth]{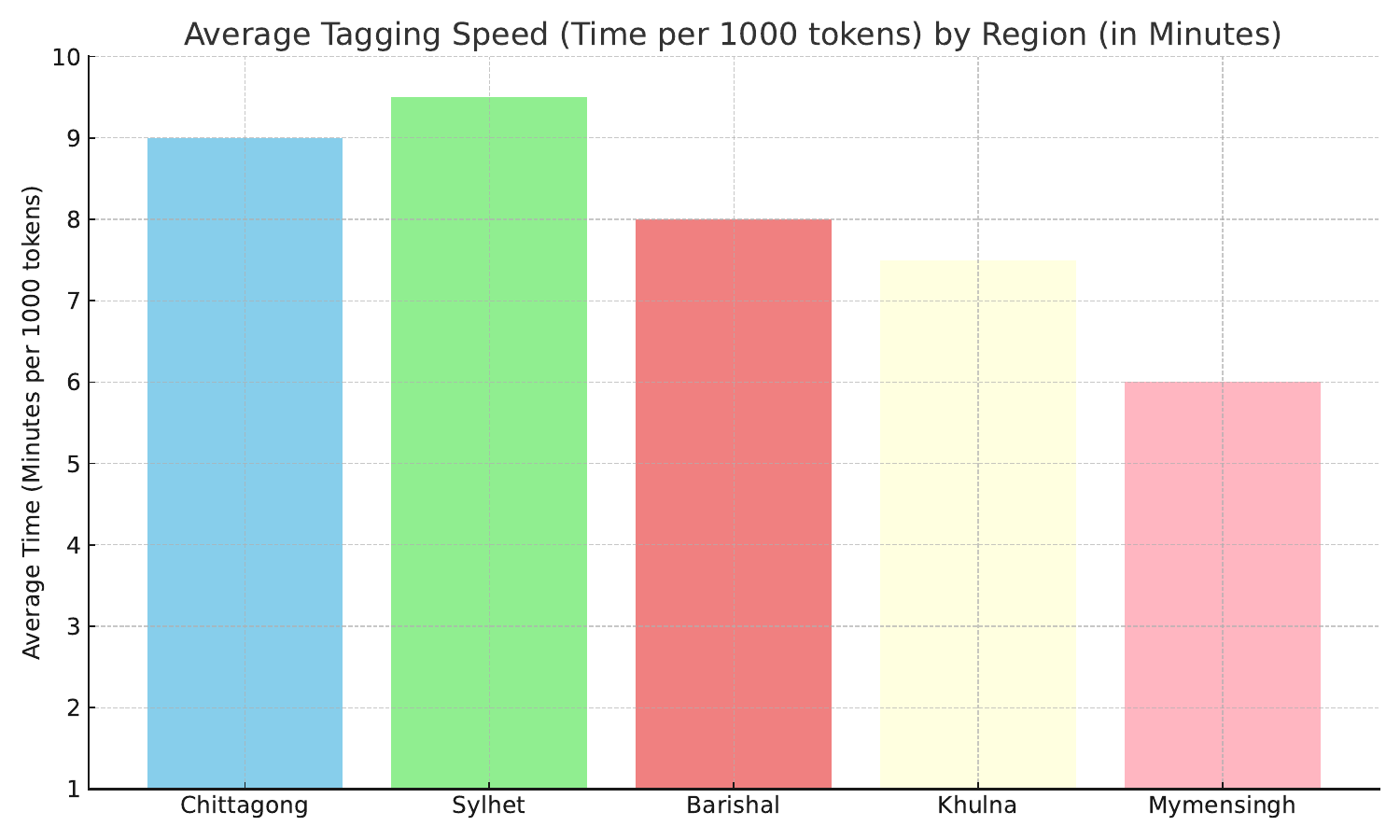} % Replace with actual image file path
        \caption{Average Tagging Speed (Time per 1000 tokens) by region in minutes.}
        \label{fig:tagging_speed}
    \end{minipage}
   
\end{figure}

The Inter-Annotator Agreement (Cohen's Kappa) graph shown in figure~\ref{fig:kappa} evaluates the consistency and reliability of the annotations performed by multiple annotators. Cohen's Kappa~\cite{kappa} is a statistical measure that assesses the level of agreement between two annotators, taking into account the possibility of agreement occurring by chance. In this study, each regional dialect was annotated by two annotators, and the Kappa scores were calculated for each pair. The graph visualizes these Kappa scores for each annotator, highlighting the level of agreement across different regional dialects. Higher Kappa scores, closer to 1, indicate a strong agreement between annotators, which is critical for ensuring the quality and consistency of the annotated data. As seen in the graph, the agreement across most regions is high, demonstrating that the annotators were well-aligned in their understanding of the tagging guidelines. This agreement is essential for the reliability of the ANCHOLIK-NER dataset, as it ensures that the entity annotations accurately represent the dialectal variations.

The Tagging Speed (Time per 1000 Tokens) by Region graph shown in figure~\ref{fig:tagging_speed} provides insights into the efficiency of the annotators when labeling entities in the dataset. The graph shows the average time each annotator took to tag 1000 tokens, grouped by region. This metric is important because it highlights the speed of annotation and reflects the annotators' familiarity with the linguistic features of their respective regional dialects. The results indicate that annotators from Mymensingh were the fastest averaging 6 min for the completion of 1000 tokens, completing the annotation in the least amount of time, while Chittagong and Sylhet annotators took slightly longer, around 9 and 9.5 mins repectively. The regions are color-coded, with Barishal taking an intermediate amount of time, followed by Khulna, which showed a similar efficiency. The graph also suggests that experience with the dialect and familiarity with the annotation process may have influenced the speed of tagging.

Following the annotation process, the dataset has been organized into separate sub-datasets for each regional dialect. The entire corpus is saved as CSV (comma-separated values) file, which is structured quite conveniently. A sample of the data structure is presented in Table~\ref{tab:dataset}, where the regions are displayed side by side for comparison purposes. Each entry in the dataset consists of:
\begin{enumerate}
    \item Sentence Number – The sentence number in the ANCHOLIK-NER dataset for each regions.
    \item Tokenized Words – Each word is treated as a separate token.
    \item Named Entity Annotations – Assigned entity tags following the BIO scheme.
\end{enumerate}

\begin{table}[htbp]
    \centering
    \caption{Dataset consists of 3 columns for each region, with the first two generated by a Python script and the third (BIO-Tags) verified by Bangla Regional Language experts.}
    \renewcommand{\arraystretch}{1.1}
    \begin{tabular}{p{2cm} p{1.5cm} p{2cm} p{2cm} p{2cm} p{2cm} p{1.5cm}}
        \toprule
        \textbf{Sentence \#} & \textbf{Sylhet} & \textbf{Chittagong} & \textbf{Barishal} & \textbf{Noakhali} & \textbf{Mymensingh} & \textbf{BIO Tags} \\
        \midrule
        1 & {\bng phuyaTay} & {\bng ephaya{I}ba} & {\bng plaUg/ga} & {\bng epalaDa} & {\bng echrhaTa} & O \\
        NaN & {\bng iselT} & {\bng iselTt/tun} & {\bng shYelt} & {\bng iselT} & {\bng iselTtn}  & B-LOC \\
        NaN & {\bng thaik} & - & {\bng idya} & {\bng eth{I}ka} & -  & O \\
        NaN &  {\bng Dhakat} & {\bng DhaHa} & {\bng dhaHa} & {\bng Dhakay} & {\bng Dhakat} & B-LOC \\
        NaN &  {\bng Aa{I}ech} & {\bng Aa{I}esY} & {\bng Aa{I}esel} & {\bng Aa{I}es} & {\bng Aa{I}es} & O \\
        \bottomrule
    \end{tabular}
    
    \label{tab:dataset}
\end{table}

The Table~\ref{tab:dataset} illustrates the variations in word structure across different regional dialects of Bangladesh for the same sentence, with specific attention to an important aspects. The presence of a "-" in the table indicates that there was no corresponding word in that dialect for the given word, highlighting a gap or difference in vocabulary between regions. For example, the word "{\bng iselT thaik}" in the Sylhet region doesn't have a match in the Chittagong region, which is represented by the "-" symbol. Two words had been merged into a single word "{\bng iselTt/tun}". This merging or splitting of words leads to discrepancies in word counts between the dialects, as seen in the varying number of words per sentence for each region. Additionally, it is important to note that there is no "-" in the actual CSV files; it has only been added in the table for comparison purposes. These differences highlight the linguistic diversity and the complex nature of regional dialects in Bangladesh.

\subsection{Outliers Checking}
Algorithm 2 was designed to ensure the quality and consistency of the annotations in the dataset. The algorithm checks for common issues in the BIO Tags column, such as outliers (tags that are incorrectly written in lowercase, like "o" instead of "O") and blank cells (missing annotations). The algorithm first processes the dataset by extracting all the unique tags and identifying any instances where tags deviate from the expected format. Specifically, it detects any outliers, such as lowercase tags or any missing tags, by checking each BIO tag against the standard format. Once any outliers or blank cells are identified, the algorithm prints the corresponding line numbers, making it easy to pinpoint where corrections are needed. The annotators are promptly notified of these discrepancies and instructed to review the flagged entries. By correcting these issues, annotators ensure that all tags are correctly formatted, and no tokens are left untagged.

\begin{algorithm}
\caption{Checking for Unique Tags and Outliers in BIO Tags}
\begin{algorithmic}[1]
\STATE \textbf{Input:} Read the data file (CSV) containing Bangla Regional sentences
\STATE tags $\leftarrow$ [ ] \COMMENT{List to store BIO Tags}
\STATE uniqueTags $\leftarrow$ [ ] \COMMENT{List to store unique BIO Tags}
\STATE lineNo $\leftarrow$ 0 \COMMENT{Line number tracker}
\FOR{each line in the data file}
    \STATE words $\leftarrow$ extract words from the row (Sentence \#, Word, BIO Tags)
    \STATE wordList $\leftarrow$ split the row into a list of words
    \IF{wordList is not empty}
        \STATE lineNo $\leftarrow$ lineNo + 1
        \STATE tags $\leftarrow$ tags + [wordList[2]] \COMMENT{Add BIO Tag to tags list}
        \IF{wordList[2] is lowercase or non-standard (e.g., "o" instead of "O")}
            \STATE PRINT "Outlier found at line: ", lineNo \COMMENT{Print line number for outlier}
        \ENDIF
        \IF{wordList[2] is empty}
            \STATE PRINT "Blank tag found at line: ", lineNo \COMMENT{Print line number for blank tag}
        \ENDIF
    \ENDIF
\ENDFOR
\FOR{each tag in tags}
    \IF{tag NOT IN uniqueTags}
        \STATE uniqueTags $\leftarrow$ uniqueTags + [tag] \COMMENT{Add unique tag to uniqueTags list}
    \ENDIF
\ENDFOR
\STATE \textbf{Output:} Print the list of uniqueTags to identify all distinct BIO Tags
\STATE \textbf{Output:} Print all the line numbers where outliers and blank tags are found
\end{algorithmic}
\end{algorithm}

\subsection{Dataset Statistics}

After the dataset was compiled, it was essential to conduct an analysis to identify both its strengths and limitations. The data analysis revealed crucial patterns and relationships within the dataset that were not initially apparent. Table~\ref{tab:data_stat} provides an overview of the dataset's content, showing that it contains 17,405 sentences, with sentence lengths ranging from 2 to 10 words. The dataset encompasses over 1 lac tokens. The analysis reveals that named entities account for only a small portion, with non-named entities making up approximately 83.4\% of the dataset, while named entities comprise 16.6\%.

\begin{table}[htbp]
    \centering
    \caption{Overview of our proposed Dataset}
    \renewcommand{\arraystretch}{1.1}
    \begin{tabular}{p{5cm}  p{4cm}}
        \toprule
        \textbf{Dataset Attributes} & \textbf{Frequency} \\
        \midrule
        Total Number of sentences & 17.405 \\
        Total Named Entities &  11,062 \\
        Total Non-Named Entities &  90,755 \\
        Sentence Length & [2-10] \\
        Entities & 10 \\
        Tagging Scheme & BIO \\
        Number of Tags & 19 \\
        \bottomrule
    \end{tabular}
    
    \label{tab:data_stat}
\end{table}

\begin{figure}[htbp]
    \centering
    % First Image - Chittagong
    \begin{minipage}{0.30\textwidth}
        \centering
        \includegraphics[width=\linewidth]{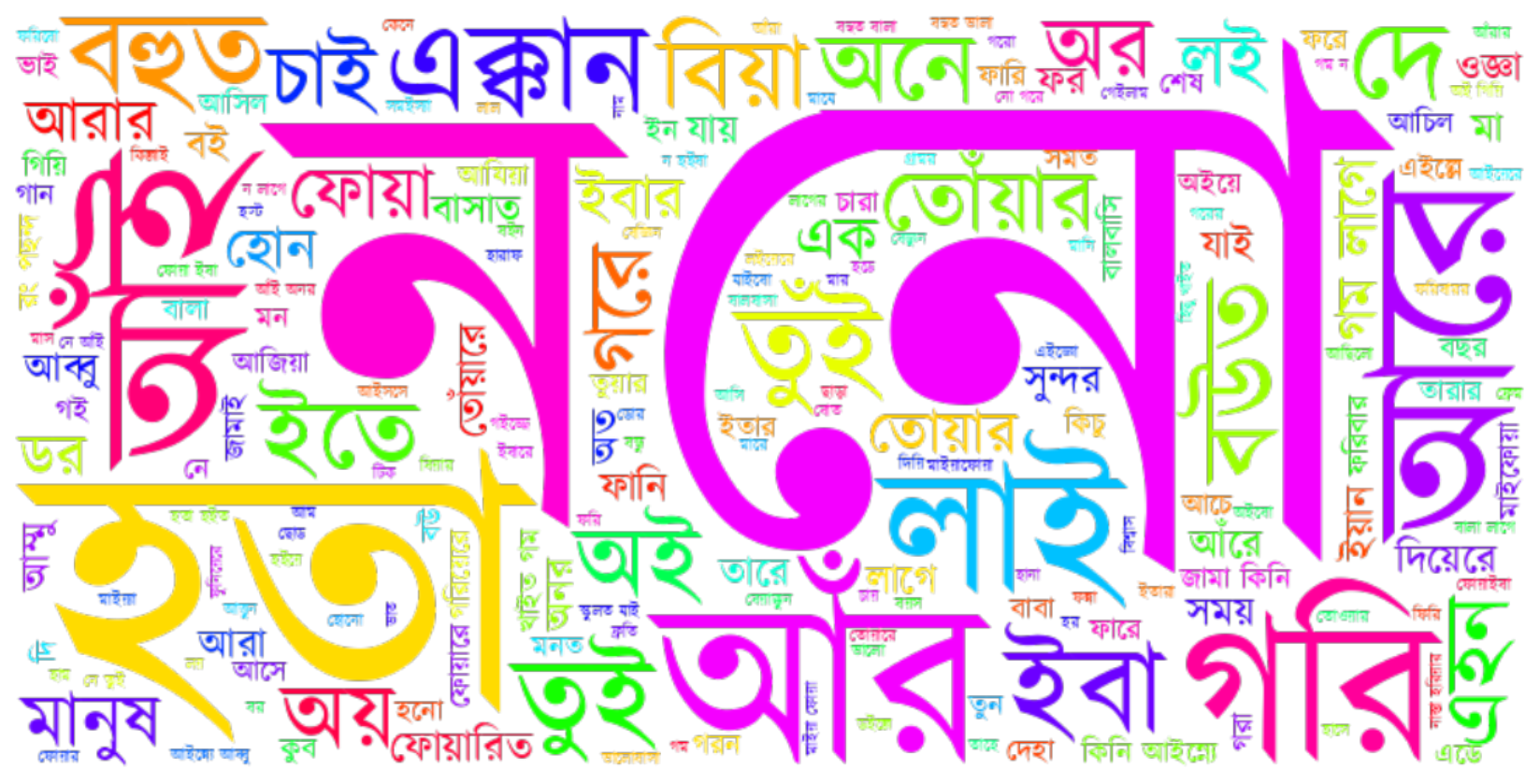} % Replace with actual image file path
        \subcaption{Chittagong}
    \end{minipage}
    \hfill
    % Second Image - Sylhet
    \begin{minipage}{0.30\textwidth}
        \centering
        \includegraphics[width=\linewidth]{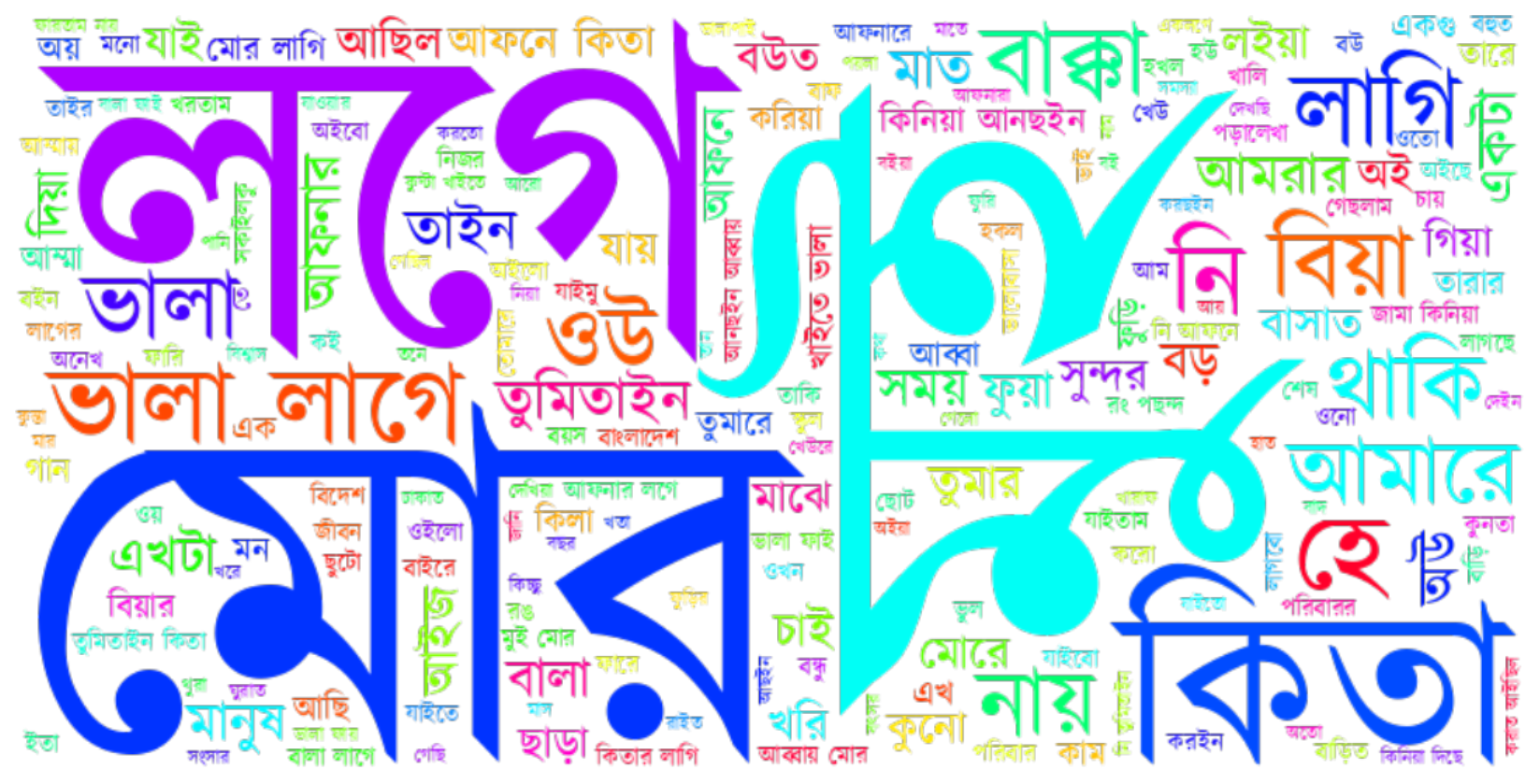} % Replace with actual image file path
        \subcaption{Sylhet}
    \end{minipage}
    \hfill
    % Third Image - Barishal
    \begin{minipage}{0.30\textwidth}
        \centering
        \includegraphics[width=\linewidth]{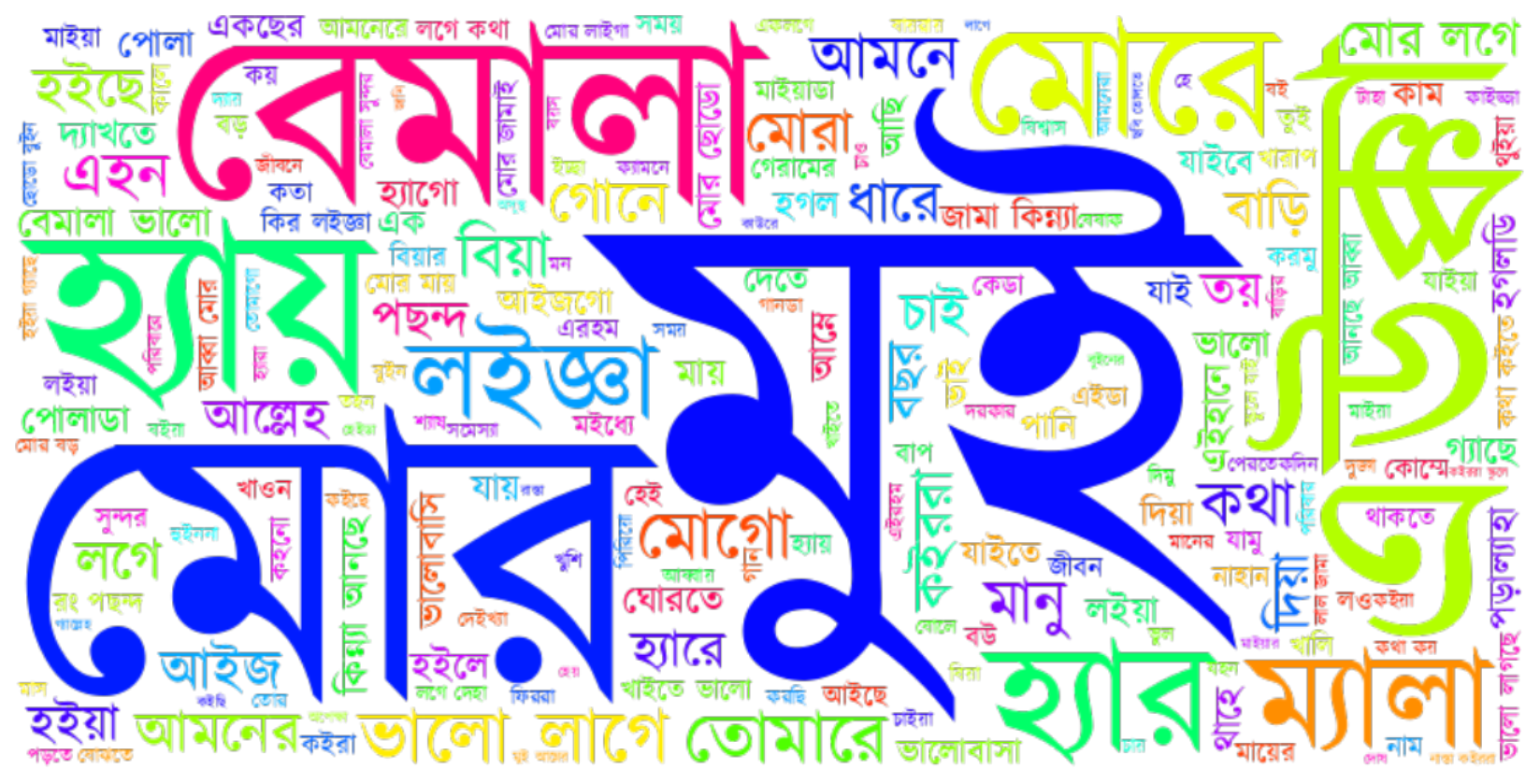} % Replace with actual image file path
        \subcaption{Barishal}
    \end{minipage}
    \vskip\baselineskip % Space between the two rows
    % Fourth and Fifth Image - Noakhali and Mymensingh (Side by Side)
    \begin{minipage}{0.30\textwidth}
        \centering
        \includegraphics[width=\linewidth]{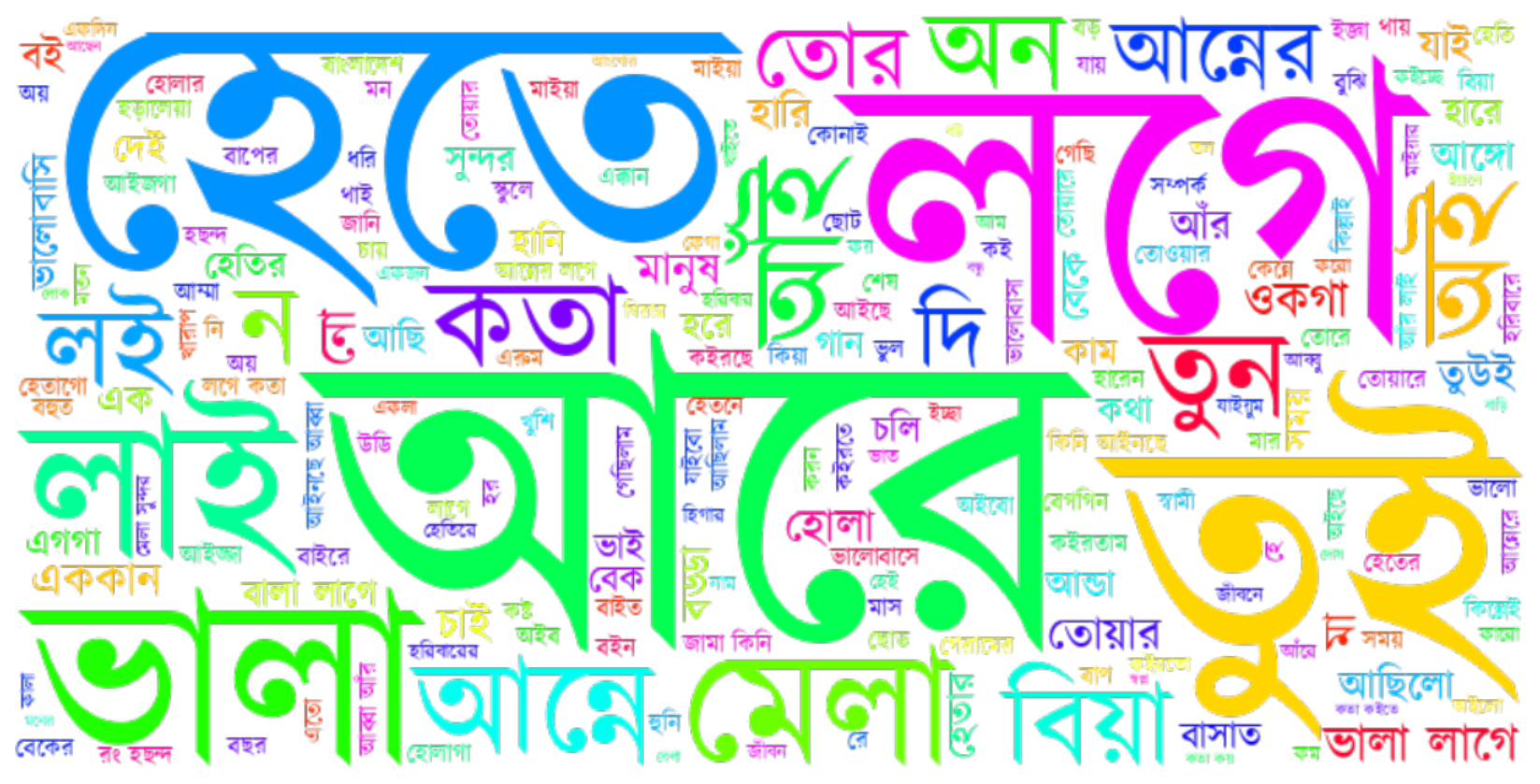} % Replace with actual image file path
        \subcaption{Noakhali}
    \end{minipage}
    \hspace{0.05\textwidth}
    \begin{minipage}{0.30\textwidth}
        \centering
        \includegraphics[width=\linewidth]{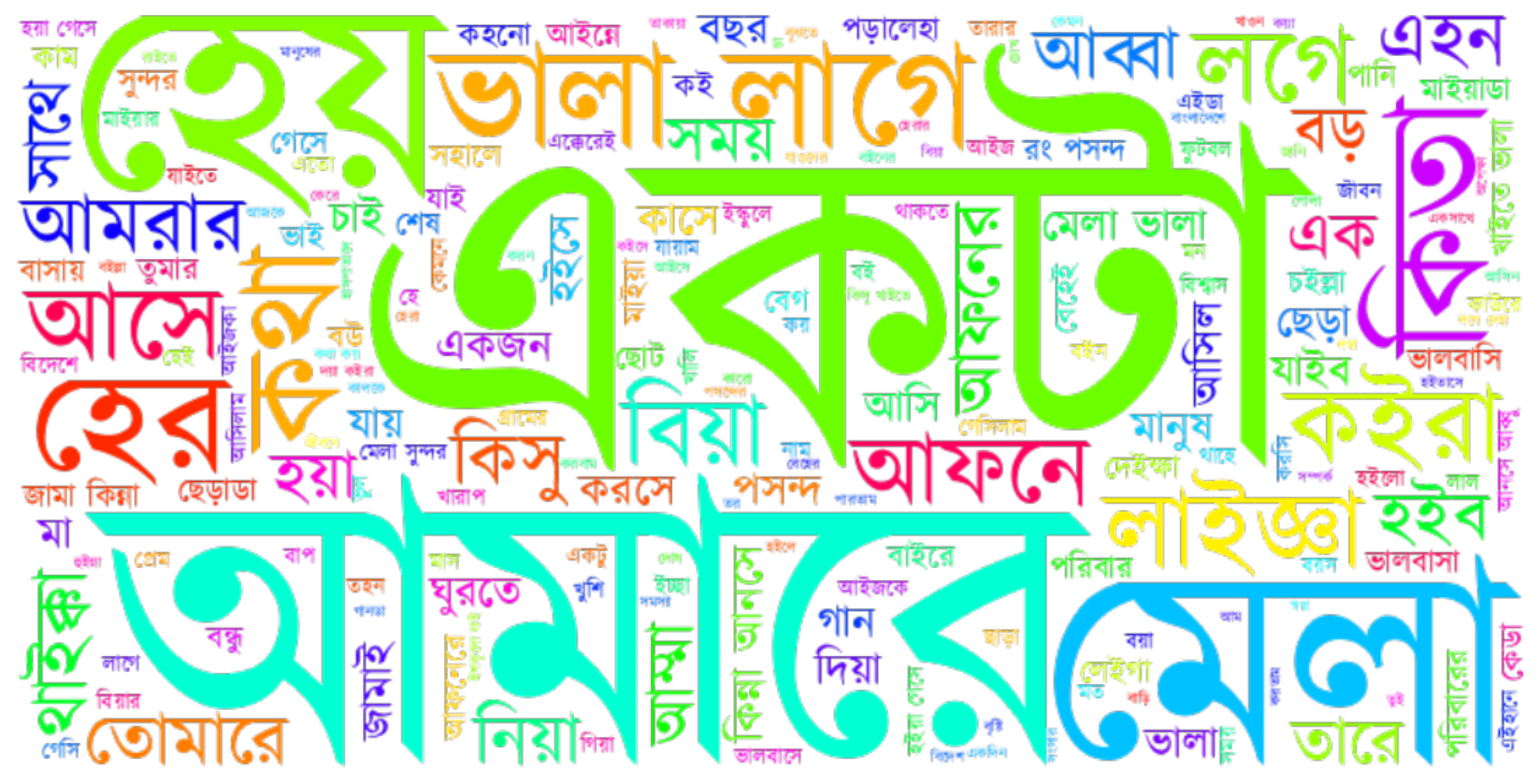} % Replace with actual image file path
        \subcaption{Mymensingh}
    \end{minipage}
    \caption{Word cloud of 5 regional Dialects}
    \label{fig:word}
\end{figure}

The word clouds presented in Figure \ref{fig:word} provide a visual representation of the most frequent terms in each of the five regional dialects: Chittagong, Sylhet, Barishal, Noakhali, and Mymensingh. These clouds highlight the unique vocabulary and linguistic features characteristic of each region, with larger words indicating higher frequency.

\begin{figure}[htbp]
    \centering
    % First Image - Chittagong
    \begin{minipage}{0.30\textwidth}
        \centering
        \includegraphics[width=\linewidth]{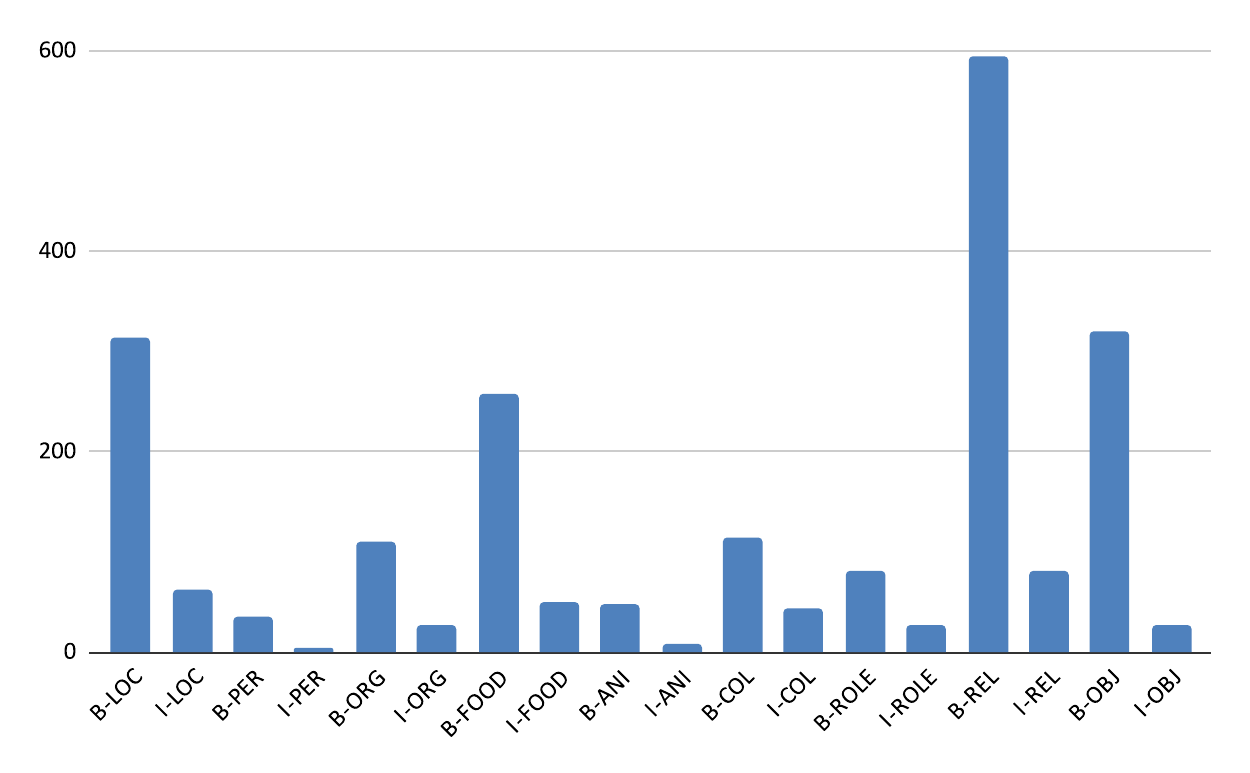} % Replace with actual image file path
        \subcaption{Chittagong}
    \end{minipage}
    \hfill
    % Second Image - Sylhet
    \begin{minipage}{0.30\textwidth}
        \centering
        \includegraphics[width=\linewidth]{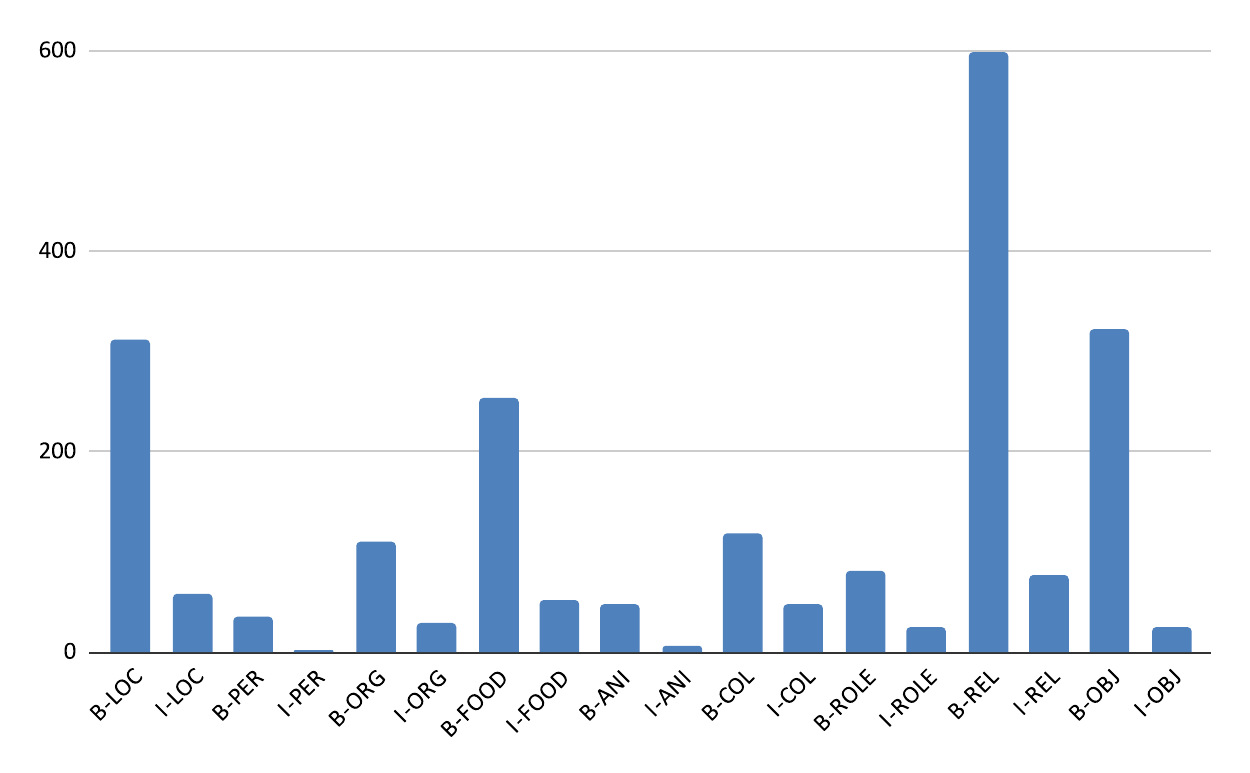} % Replace with actual image file path
        \subcaption{Sylhet}
    \end{minipage}
    \hfill % Space between the two rows
    % Third Image - Barishal
    \begin{minipage}{0.30\textwidth}
        \centering
        \includegraphics[width=\linewidth]{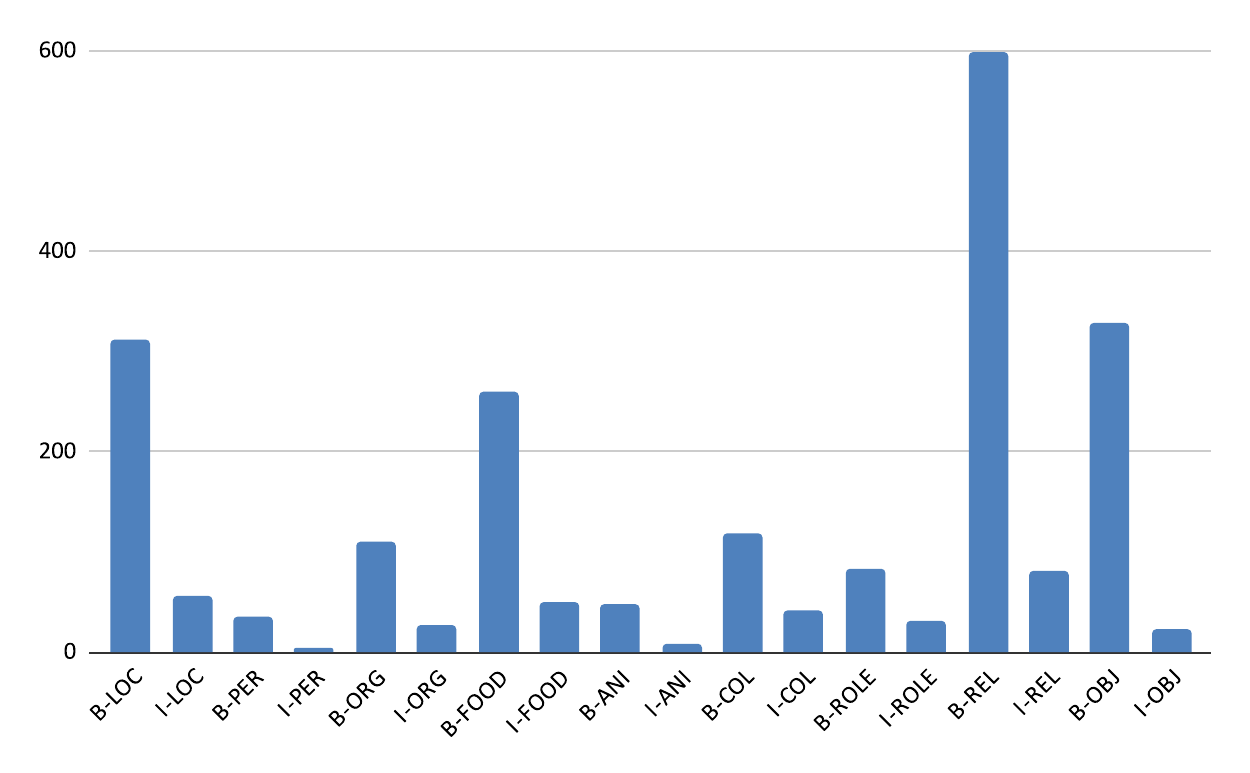} % Replace with actual image file path
        \subcaption{Barishal}
    \end{minipage}
    \vskip\baselineskip % Space between the two rows
    % Fourth and Fifth Image - Noakhali and Mymensingh (Side by Side)
    \begin{minipage}{0.30\textwidth}
        \centering
        \includegraphics[width=\linewidth]{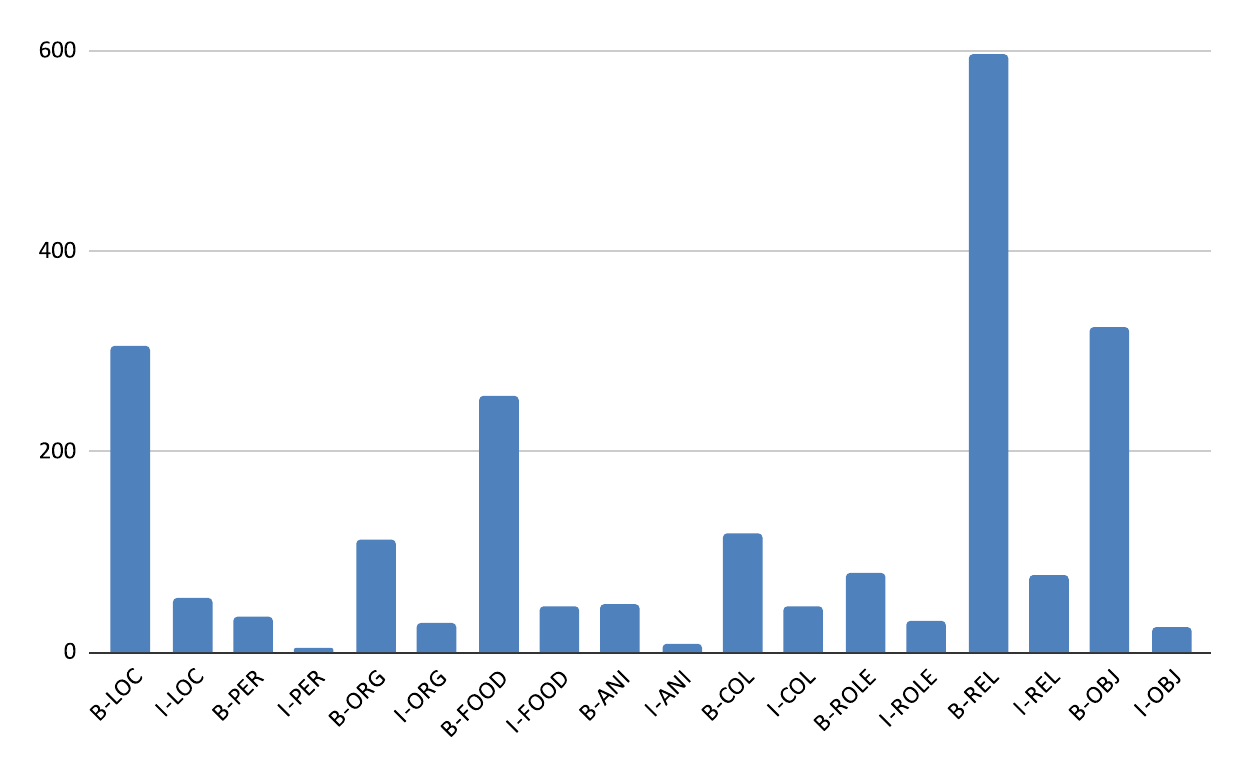} % Replace with actual image file path
        \subcaption{Noakhali}
    \end{minipage}
    \hspace{0.05\textwidth} % Space between the two rows
    \begin{minipage}{0.30\textwidth}
        \centering
        \includegraphics[width=\linewidth]{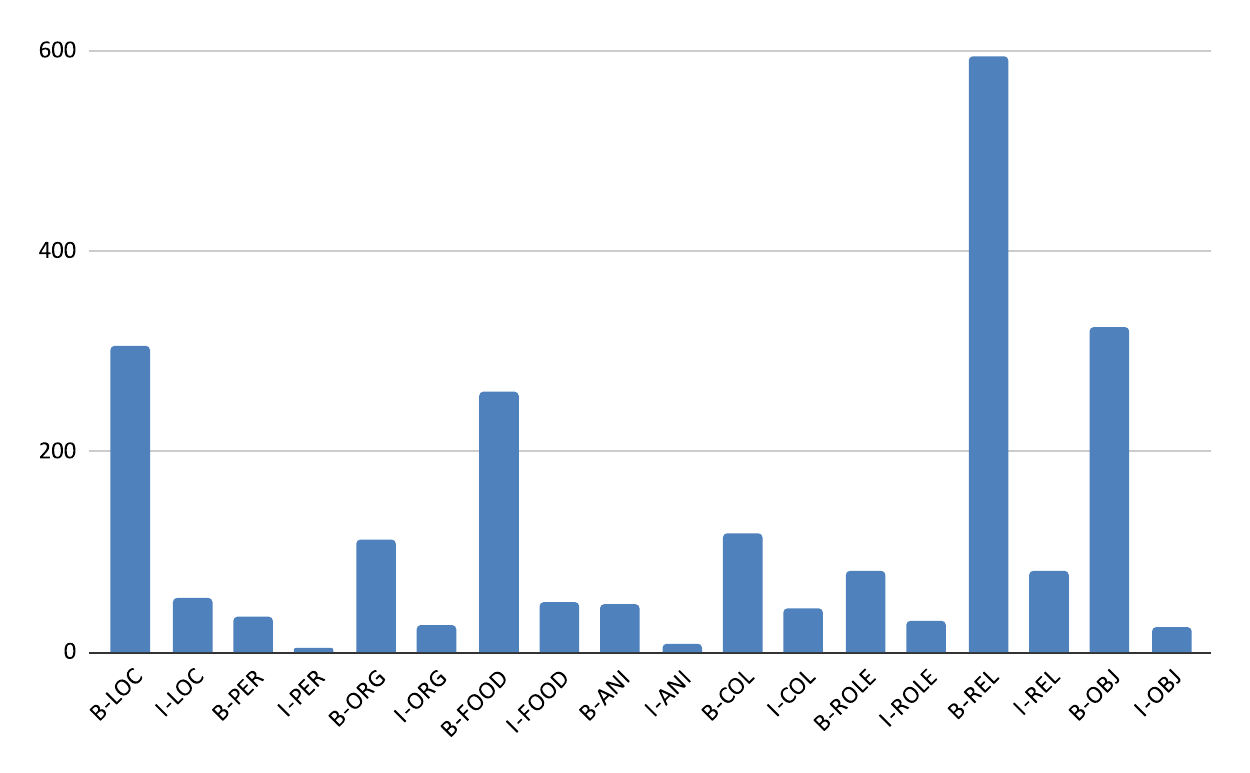} % Replace with actual image file path
        \subcaption{Mymensingh}
    \end{minipage}
    \caption{Frequency of Named Entities for 5 regional Dialects}
    \label{fig:freq}
\end{figure}

Table~\ref{tab:sample1} and Figure~\ref{fig:freq} provide a detailed breakdown of the distribution and frequency of named entity types across the five regional dialects in the dataset. Table~\ref{tab:sample1} lists the total instances of each named entity type for Barishal, Sylhet, Chittagong, Noakhali, and Mymensingh. The frequency distribution of these entities is further visualized in Figure~\ref{fig:freq}, where each subfigure (a-e) represents the frequency of named entities in the corresponding region.

\begin{table}[htpb]
    \centering
    \caption{Total instances of Named Entity Types in five Regions}
    \renewcommand{\arraystretch}{1.1} % Adjust row height
    \begin{tabular}{p{3cm} r r r r r r}
        \toprule
        \textbf{Named Entity Type} & \textbf{Barishal} & \textbf{Sylhet} & \textbf{Chittagong} & \textbf{Noakhali} & \textbf{Mymensingh} & \textbf{Total Instances} \\
        \midrule
        Person (PER) & 39 & 38 & 39 & 39 & 39 & 194 \\
        Location (LOC) & 369 & 371 & 377 & 361 & 362 & 1840 \\
        Organization (ORG) & 139 & 141 & 139 & 141 & 140 & 700 \\
        Food (FOOD) & 310 & 308 & 308 & 303 & 312 & 1541 \\
        Animal (ANI) & 57 & 56 & 57 & 57 & 57 & 284 \\
        Colour (COL) & 162 & 167 & 160 & 164 & 163 & 816 \\
        Role (ROLE) & 114 & 107 & 109 & 111 & 113 & 554 \\
        Relation (REL) & 681 & 677 & 676 & 676 & 676 & 3386 \\
        Object (OBJ) & 352 & 348 & 348 & 350 & 349 & 1747 \\
        Miscellaneous (O) & 17928 & 18750 & 18177 & 17957 & 17943 & 90,755 \\
        \bottomrule
    \end{tabular}
    
    \label{tab:sample1}
\end{table}

\subsection{Data Distribution}
The dataset was split into 80\% for training and 20\% for testing, based on the total number of sentences. This distribution was done for each of the five regions. This 80:20 split ensures that the models are trained on a large portion of the data while being tested on a smaller, distinct subset for unbiased evaluation.

\subsection{Data Availability}
\label{Data_Availability}

The ANCHOLIK-NER dataset, which contains annotated named entities in Bangla regional dialects, is publicly available for research and academic purposes. Researchers interested in utilizing the dataset can access it through the following link:\href{https://data.mendeley.com/datasets/gbkszkt8z3/3}{Dataset of Named Entity Recognition for Regional Bangla Language (Original data)} (Mendeley Data)

% Assuming that a sentence consists of \( n \) number of words. If the sentence is represented as a set \( s = \{ w_1, w_2, w_3, \dots, w_n \} \), where each word is considered as a token, the required output is such that \( t = \{ t_1, t_2, t_3, \dots, t_n \} \). Here, \( t_i \in \{ \textit{\text{B-PER}, \text{I-PER}, \text{B-LOC}, \text{I-LOC}, \text{B-ORG}, \text{I-ORG}, \text{B-REL}, \text{I-REL}, \text{B-FOOD}, \text{I-FOOD}, \text{B-ANI}, \text{I-ANI}, O \}} \). The context of the sentence is required to be considered when tagging each token present in that sentence.

\section{Methodology}
\label{Methodology}

\begin{figure}[htbp]
    \centering
    \includegraphics[width=0.7\linewidth]{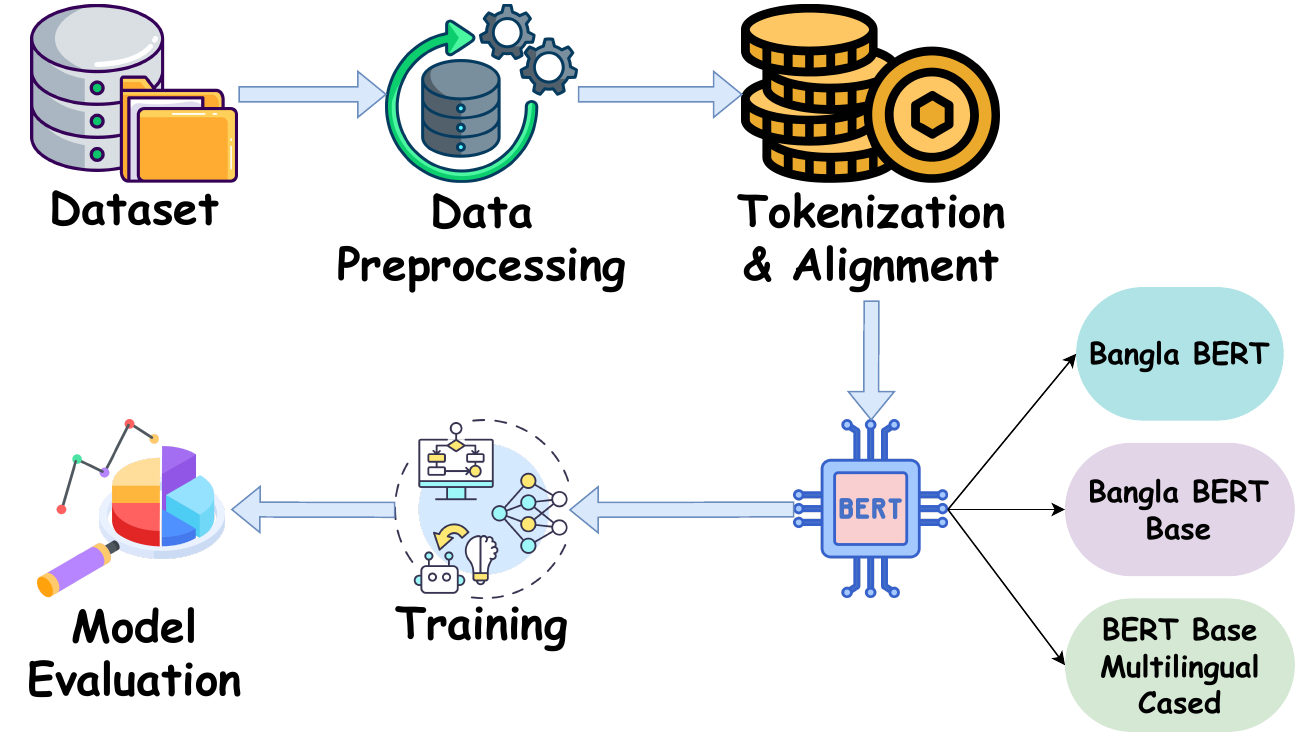}
    \caption{Methodology}
    \label{fig:methodology}
\end{figure}

In this section, we focus on developing a Named Entity Recognition (NER) model for Bangla regional dialects. Following the dataset preparation, preprocessing, and tokenization steps, the methodology emphasizes the use of specialized and multilingual BERT models for NER tasks. Figure~\ref{fig:methodology} gives an overview of our proposed methodology.

\subsection{Models Overview}
Named Entity Recognition (NER) in Bangla requires models that are specifically fine-tuned to understand the language's unique syntax, morphology, and regional variations. Given the complexity of Bangla and its regional dialects, transformer-based models like Bangla BERT, Bangla BERT Base, and BERT Base Multilingual Cased have shown great promise in achieving high accuracy for NER tasks. 

\subsubsection{Bangla Bert}
Bangla BERT~\cite{banglabert} is pre-trained specifically on Bangla data and fine-tuned for a range of Bangla NLP tasks, including NER. This model is particularly useful in NER tasks as it is fine-tuned to capture the syntactic and semantic intricacies of the Bangla language, ensuring high accuracy in identifying named entities. The choice of Bangla BERT ensures that the model is fine-tuned to the specific linguistic structure of Bangla, including its morphology and syntactic dependencies.

\subsubsection{Bangla Bert Base}
The Bangla BERT Base~\cite{Sagor_2020} variant is based on the original BERT model architecture but pre-trained on a large corpus of Bangla text. It is well-suited for NER tasks due to its balanced structure, which optimizes performance while maintaining computational efficiency, especially when handling complex dialectal variations in regional Bangla. This variant is essential for situations where model performance needs to be optimized while managing computational resources. By using the BERT Base architecture, which has been proven to work effectively on various NLP tasks, we aim to achieve an optimal balance between computational cost and performance.

\subsubsection{BERT Base Multilingual Cased}
The BERT Base Multilingual Cased model~\cite{Bertbase} is trained on a massive multilingual corpus, supporting over 100 languages, including Bangla. The model uses the transformer architecture to understand text in multiple languages simultaneously, making it highly useful for cross-linguistic tasks such as NER. This multilingual model was selected to evaluate how well a general-purpose language model can adapt to the specific requirements of Bangla regional dialects. It is particularly useful for understanding language diversity and cross-linguistic transfer in NLP tasks.

\subsection{Evaluation Metrics}
To assess the performance of the NER models, we use Precision, Recall, and F1-Score as our primary evaluation metrics. These metrics were selected because they provide a more balanced and informative evaluation of the model's performance, especially in tasks like NER, which often involve imbalanced datasets.

\subsubsection{Precision}
Precision measures the proportion of true positive named entities identified by the model out of all the named entities that the model identified as positive. A high precision score indicates that the model is making fewer false positive errors, meaning it correctly identifies relevant named entities without mistakenly classifying non-entities as entities. Precision is important when it is crucial to ensure that the entities detected by the model are relevant and correctly identified. Equation 1 displays the formula of precision.

\begin{equation}
\text{Precision} = \frac{TP}{TP + FP}
\end{equation}

Where:
\begin{itemize}
    \item \( TP \) = True Positives
    \item \( FP \) = False Positives
\end{itemize}

\subsubsection{Recall}
Recall measures the proportion of true positive named entities identified by the model out of all the actual positive named entities in the dataset. A high recall score indicates that the model is capturing most of the true named entities and not missing important information. This metric is crucial for tasks where it is important to minimize false negatives, especially when the named entities are important for the overall understanding of the text. Equation 2 displays the formula of recall.

\begin{equation}
\text{Recall} = \frac{TP}{TP + FN}
\end{equation}

Where:
\begin{itemize}
    \item \( TP \) = True Positives
    \item \( FN \) = False Negatives
\end{itemize}

\subsubsection{F1-Score}
The F1-Score is the harmonic mean of precision and recall, and it is particularly useful when dealing with imbalanced datasets. The F1-score provides a single metric that balances both precision and recall, ensuring that the model does not overemphasize one metric at the expense of the other. This metric is highly effective for NER tasks, where the correct identification of entities is critical, and both false positives and false negatives should be minimized~\cite{f1}. In our NER tasks, especially for Bangla regional dialects, an imbalance in named entities (e.g., fewer occurrences of specific entity types) is common, making the F1-score an essential metric for evaluating the model’s overall performance. By considering both precision and recall, the F1-score ensures that the model achieves a balance between identifying all relevant entities while avoiding the misclassification of non-entities as named entities. Equation 3 displays the formula for the F1-Score. Equation 3 displays the formula of F1-Score.

\begin{equation}
\text{F1-Score} = 2 \times \frac{\text{Precision} \times \text{Recall}}{\text{Precision} + \text{Recall}}
\end{equation}

\begin{table}[htbp]
\centering
\caption{Performance of Bangla BERT}
\label{tab:Bangla_BERT}
\renewcommand{\arraystretch}{1.25}
\begin{tabular}{|c|c|c c c c c|}
\toprule
\textbf{Model} & \textbf{Batch Size} & \textbf{Region} & \textbf{Epoch} & \textbf{Precision} & \textbf{Recall} & \textbf{F1-score} \\
\midrule
\multirow{40}{*}{Bangla BERT} & \multirow{20}{*}{8} & \multirow{4}{*}{Barishal} & 05 & 0.69931 & 0.73465 & 0.71654 \\
 &  &  & 10 & 0.77915 & 0.79602 & 0.78750 \\
 &  &  & 15 & 0.80900 & 0.81046 & 0.80973 \\
 &  &  & 20 & 0.83650 & 0.79422 & \textbf{0.81481} \\ \cline{3-3} \cline{4-7}
 &  & \multirow{4}{*}{Chittagong} & 05 & 0.66775 & 0.65806 & 0.66287 \\
 &  &  & 10 & 0.76334 & 0.69193 & 0.72588 \\
 &  &  & 15 & 0.76329 & 0.71774 & 0.73981 \\
 &  &  & 20 & 0.76627 & 0.74032 & \textbf{0.75307} \\ \cline{3-3} \cline{4-7}
 &  & \multirow{4}{*}{Mymensingh} & 05 & 0.74353 & 0.76639 & 0.75479 \\
 &  &  & 10 & 0.79446 & 0.82377 & 0.80885 \\
 &  &  & 15 & 0.80360 & 0.82172 & 0.81256 \\
 &  &  & 20 & 0.82780 & 0.81762 & \textbf{0.82268} \\ \cline{3-3} \cline{4-7}
 &  & \multirow{4}{*}{Noakhali} & 05 & 0.65669 & 0.69459 & 0.67511 \\
 &  &  & 10 & 0.78694 & 0.76350 & 0.77504 \\
 &  &  & 15 & 0.77692 & 0.75232 & 0.76442 \\
 &  &  & 20 & 0.79166 & 0.77839 & \textbf{0.78497} \\ \cline{3-3} \cline{4-7}
 &  & \multirow{4}{*}{Sylhet} & 05 & 0.75412 & 0.74591 & 0.75000 \\
 &  &  & 10 & 0.76363 & 0.76225 & 0.76294 \\
 &  &  & 15 & 0.81106 & 0.77132 & \textbf{0.79069} \\
 &  &  & 20 & 0.79482 & 0.78039 & 0.78754 \\ \cline{2-3} \cline{4-7}
 & \multirow{20}{*}{16} & \multirow{4}{*}{Barishal} & 05 & 0.66725 & 0.67689 & 0.67204 \\
 &  &  & 10 & 0.72972 & 0.77978 & 0.75392 \\
 &  &  & 15 & 0.77640 & 0.79602 & 0.78609 \\
 &  &  & 20 & 0.78368 & 0.79783 & \textbf{0.79069} \\ \cline{3-3} \cline{4-7}
 &  & \multirow{4}{*}{Chittagong} & 05 & 0.60202 & 0.57580 & 0.58862 \\
 &  &  & 10 & 0.66944 & 0.64677 & 0.65791 \\
 &  &  & 15 & 0.73445 & 0.70483 & 0.71934 \\
 &  &  & 20 & 0.75043 & 0.69838 & \textbf{0.72347} \\ \cline{3-3} \cline{4-7}
 &  & \multirow{4}{*}{Mymensingh} & 05 & 0.66358 & 0.73565 & 0.69776 \\
 &  &  & 10 & 0.77354 & 0.79098 & 0.78216 \\
 &  &  & 15 & 0.81069 & 0.80737 & 0.80903 \\
 &  &  & 20 & 0.82244 & 0.82581 & \textbf{0.82413} \\ \cline{3-3} \cline{4-7}
 &  & \multirow{4}{*}{Noakhali} & 05 & 0.56856 & 0.63314 & 0.59911 \\
 &  &  & 10 & 0.72413 & 0.74301 & 0.73345 \\
 &  &  & 15 & 0.77042 & 0.73743 & 0.75356 \\
 &  &  & 20 & 0.77716 & 0.78584 & \textbf{0.78148} \\ \cline{3-3} \cline{4-7}
 &  & \multirow{4}{*}{Sylhet} & 05 & 0.65277 & 0.68239 & 0.66725 \\
 &  &  & 10 & 0.75591 & 0.75317 & 0.75454 \\
 &  &  & 15 & 0.77614 & 0.76769 & 0.77189 \\
 &  &  & 20 & 0.79777 & 0.78039 & \textbf{0.78899} \\
\bottomrule
\end{tabular}
\end{table}

\begin{table}[htbp]
\centering
\caption{Performance of Bangla BERT Base}
\label{tab:Bangla_BERT_Base}
\renewcommand{\arraystretch}{1.25}
\begin{tabular}{|c|c|c c c c c|}
\toprule
\textbf{Model} & \textbf{Batch Size} & \textbf{Region} & \textbf{Epoch} & \textbf{Precision} & \textbf{Recall} & \textbf{F1-score} \\
\midrule
\multirow{40}{*}{\begin{tabular}[c]{@{}c@{}}Bangla BERT\\ Base\end{tabular}} & \multirow{20}{*}{08} & \multirow{4}{*}{Barishal} & 05 & 0.80711 & 0.76424 & 0.78509 \\ 
 &  &  & 10 & 0.81292 & 0.76554 & 0.78852 \\
 &  &  & 15 & 0.82216 & 0.77849 & \textbf{0.79973} \\
 &  &  & 20 & 0.82753 & 0.77072 & 0.79812 \\ \cline{3-3} \cline{4-7} 
 &  & \multirow{4}{*}{Chittagong} & 05 & 0.76062 & 0.64314 & 0.69696 \\ 
 &  &  & 10 & 0.75888 & 0.65379 & 0.70243 \\ 
 &  &  & 15 & 0.77864 & 0.66045 & \textbf{0.71469} \\ 
 &  &  & 20 & 0.78964 & 0.64980 & 0.71292 \\ \cline{3-3} \cline{4-7} 
 &  & \multirow{4}{*}{Mymensingh} & 05 & 0.79663 & 0.77645 & 0.78641 \\ 
 &  &  & 10 & 0.82942 & 0.78986 & \textbf{0.80916} \\ 
 &  &  & 15 & 0.81720 & 0.79284 & 0.80484 \\ 
 &  &  & 20 & 0.82453 & 0.79135 & 0.80760 \\ \cline{3-3} \cline{4-7} 
 &  & \multirow{4}{*}{Noakhali} & 05 & 0.80613 & 0.73382 & 0.76828 \\ 
 &  &  & 10 & 0.82013 & 0.73088 & 0.77293 \\
 &  &  & 15 & 0.81892 & 0.73823 & 0.77648 \\ 
 &  &  & 20 & 0.81451 & 0.74264 & \textbf{0.77692} \\ \cline{3-3} \cline{4-7} 
 &  & \multirow{4}{*}{Sylhet} & 05 & 0.81481 & 0.70375 & 0.75522 \\ 
 &  &  & 10 & 0.80821 & 0.73852 & 0.77180 \\ 
 &  &  & 15 & 0.84025 & 0.73157 & \textbf{0.78215} \\ 
 &  &  & 20 & 0.82884 & 0.72739 & 0.77481 \\ \cline{2-3} \cline{4-7} 
 & \multirow{20}{*}{16} & \multirow{4}{*}{Barishal} & 05 & 0.78035 & 0.74093 & 0.76013 \\ 
 &  &  & 10 & 0.79782 & 0.76165 & 0.77932 \\ 
 &  &  & 15 & 0.80636 & 0.75518 & \textbf{0.77993} \\ 
 &  &  & 20 & 0.82419 & 0.75906 & 0.79028 \\ \cline{3-3} \cline{4-7} 
 &  & \multirow{4}{*}{Chittagong} & 05 & 0.74720 & 0.62183 & 0.67877 \\ 
 &  &  & 10 & 0.76443 & 0.65246 & 0.70402 \\ 
 &  &  & 15 & 0.77974 & 0.64580 & 0.70648 \\ 
 &  &  & 20 & 0.78378 & 0.65645 & \textbf{0.71449} \\ \cline{3-3} \cline{4-7} 
 &  & \multirow{4}{*}{Mymensingh} & 05 & 0.79750 & 0.76304 & 0.77989 \\ 
 &  &  & 10 & 0.81933 & 0.77049 & 0.79416 \\ 
 &  &  & 15 & 0.81212 & 0.79880 & \textbf{0.80540} \\ 
 &  &  & 20 & 0.83360 & 0.76900 & 0.80000 \\ \cline{3-3} \cline{4-7} 
 &  & \multirow{4}{*}{Noakhali} & 05 & 0.77777 & 0.71029 & 0.74250 \\ 
 &  &  & 10 & 0.80913 & 0.72941 & 0.76720 \\ 
 &  &  & 15 & 0.78615 & 0.75147 & 0.76842 \\ 
 &  &  & 20 & 0.80868 & 0.73970 & \textbf{0.77265} \\ \cline{3-3} \cline{4-7} 
 &  & \multirow{4}{*}{Sylhet} & 05 & 0.79047 & 0.69262 & 0.73832 \\ 
 &  &  & 10 & 0.82108 & 0.71488 & 0.76431 \\ 
 &  &  & 15 & 0.82103 & 0.72739 & \textbf{0.77138} \\ 
 &  &  & 20 & 0.82467 & 0.70653 & 0.76104 \\ \bottomrule
\end{tabular}%
\end{table}

\begin{table}[htbp]
\centering
\caption{Performance of BERT Base Multilingual Cased}
\label{tab:BERT_Base_Multilingual_Cased}
\renewcommand{\arraystretch}{1.25}
\begin{tabular}{|c|c|c c c c c|}
\toprule
\textbf{Model} & \textbf{Batch Size} & \textbf{Region} & \textbf{Epoch} & \textbf{Precision} & \textbf{Recall} & \textbf{F1-score} \\
\midrule
\multirow{40}{*}{\begin{tabular}[c]{@{}c@{}}BERT Base \\ Multilingual Cased\end{tabular}} & \multirow{20}{*}{08} & \multirow{4}{*}{Barishal} & 05 & 0.72710 & 0.72448 & 0.72579 \\  
 &  &  & 10 & 0.77556 & 0.77416 & 0.77486 \\ 
 &  &  & 15 & 0.78690 & 0.77055 & \textbf{0.77863} \\ 
 &  &  & 20 & 0.78451 & 0.75971 & 0.77191 \\ \cline{3-3} \cline{4-7} 
 &  & \multirow{4}{*}{Chittagong} & 05 & 0.74456 & 0.70801 & 0.72582 \\ 
 &  &  & 10 & 0.79777 & 0.74074 & 0.73820 \\ 
 &  &  & 15 & 0.77390 & 0.74590 & 0.75964 \\ 
 &  &  & 20 & 0.77531 & 0.75258 & \textbf{0.76377} \\ \cline{3-3} \cline{4-7} 
 &  & \multirow{4}{*}{Mymensingh} & 05 & 0.78199 & 0.78476 & 0.78337 \\ 
 &  &  & 10 & 0.82411 & 0.80513 & 0.81451 \\ 
 &  &  & 15 & 0.83455 & 0.80425 & 0.81912 \\ 
 &  &  & 20 & 0.85682 & 0.79752 & \textbf{0.82611} \\ \cline{3-3} \cline{4-7} 
 &  & \multirow{4}{*}{Noakhali} & 05 & 0.73551 & 0.76977 & 0.75225 \\ 
 &  &  & 10 & 0.80344 & 0.77943 & 0.79125 \\ 
 &  &  & 15 & 0.79355 & 0.80052 & 0.79702 \\ 
 &  &  & 20 & 0.81914 & 0.81195 & \textbf{0.81553} \\ \cline{3-3} \cline{4-7} 
 &  & \multirow{4}{*}{Sylhet} & 05 & 0.74363 & 0.73100 & 0.73726 \\ 
 &  &  & 10 & 0.80471 & 0.76228 & 0.78292 \\ 
 &  &  & 15 & 0.84789 & 0.77211 & 0.80823 \\ 
 &  &  & 20 & 0.82053 & 0.82578 & \textbf{0.82315} \\ \cline{2-3} \cline{4-7} 
 & \multirow{20}{*}{16} & \multirow{4}{*}{Barishal} & 05 & 0.70953 & 0.73261 & 0.72088 \\  
 &  &  & 10 & 0.75559 & 0.73170 & 0.74346 \\  
 &  &  & 15 & 0.77674 & 0.75429 & 0.76535 \\  
 &  &  & 20 & 0.79061 & 0.76061 & \textbf{0.77532} \\ \cline{3-3} \cline{4-7} 
 &  & \multirow{4}{*}{Chittagong} & 05 & 0.69299 & 0.66494 & 0.67868 \\  
 &  &  & 10 & 0.75837 & 0.72179 & 0.73962 \\ 
 &  &  & 15 & 0.78317 & 0.72179 & 0.75123 \\ 
 &  &  & 20 & 0.79329 & 0.73385 & \textbf{0.76241} \\ \cline{3-3} \cline{4-7} 
 &  & \multirow{4}{*}{Mymensingh} & 05 & 0.74825 & 0.75819 & 0.75318 \\ 
 &  &  & 10 & 0.79464 & 0.78830 & 0.79146 \\ 
 &  &  & 15 & 0.82384 & 0.80779 & \textbf{0.81574} \\  
 &  &  & 20 & 0.81478 & 0.81045 & 0.81261 \\ \cline{3-3} \cline{4-7} 
 &  & \multirow{4}{*}{Noakhali} & 05 & 0.70320 & 0.73286 & 0.71772 \\ 
 &  &  & 10 & 0.78892 & 0.80140 & 0.79511 \\ 
 &  &  & 15 & 0.78747 & 0.78471 & 0.78609 \\ 
 &  &  & 20 & 0.80926 & 0.78295 & \textbf{0.79589} \\ \cline{3-3} \cline{4-7} 
 &  & \multirow{4}{*}{Sylhet} & 05 & 0.68643 & 0.68275 & 0.68458 \\ 
 &  &  & 10 & 0.78133 & 0.76318 & 0.77215 \\ 
 &  &  & 15 & 0.80461 & 0.74709 & 0.77479 \\ 
 &  &  & 20 & 0.79390 & 0.76764 & \textbf{0.78055} \\ \bottomrule
\end{tabular}
\end{table}

\section{Result Analysis}
\label{Result_Analysis}

The performance of three different BERT models—Bangla BERT, Bangla BERT Base, and BERT Base Multilingual Cased—was evaluated for Named Entity Recognition (NER) across five regional dialects of Bangla: Barishal, Chittagong, Mymensingh, Noakhali, and Sylhet. The models were trained with a particular learning rates (2e-5), different batch sizes (8, 16) and epochs(5, 10, 15, 20). Their performance was assessed based on precision, recall, and F1-score.

The Bangla BERT model (Table \ref{tab:Bangla_BERT}) was evaluated across five Bangla regional dialects with varying epochs and a learning rate with batch sizes of 8 and 16. The results show that Bangla BERT performed best in the Mymensingh region, achieving the highest F1-score of \textbf{82.268\%} at 
\textbf{epoch 20}. In Barishal, it also performed well, reaching an F1-score of \textbf{81.481\%} at \textbf{epoch 20}. Sylhet and Noakhali showed moderate performance, with Sylhet achieving a peak F1-score of \textbf{78.754\%} at \textbf{epoch 20}, and Noakhali reaching \textbf{78.497\%} at \textbf{epoch 20}. The Chittagong region, however, showed relatively lower performance compared to the other regions, with the highest F1-score of \textbf{75.307\%} at \textbf{epoch 20}. Overall, Bangla BERT demonstrated strong performance, with its highest F1-scores observed in Mymensingh and Barishal.

The Bangla BERT Base model (Table \ref{tab:Bangla_BERT_Base})  performed well across all regions, with the highest F1-score of \textbf{79.973\%} in Barishal at \textbf{epoch 15} and \textbf{80.916\%} in Mymensingh at \textbf{epoch 10}, demonstrating its ability to recognize named entities effectively. In Chittagong, the model achieved a peak F1-score of \textbf{71.469\%} at \textbf{epoch 15}, while Sylhet saw its highest F1-score of \textbf{78.215\%} at \textbf{epoch 15}. Noakhali exhibited a consistent improvement in performance, with a peak F1-score of \textbf{77.692\%} at \textbf{epoch 20}. Overall, Bangla BERT Base showed solid performance across most regions, with particularly notable results in Mymensingh and Sylhet, while regions like Chittagong and Noakhali saw more modest improvements. These results suggest that the model performs well but could benefit from further fine-tuning for dialects with more complex linguistic features.

The BERT Base Multilingual Cased model (Table \ref{tab:BERT_Base_Multilingual_Cased}) showed strong performance across most regions, achieving the highest F1-scores in Mymensingh and Sylhet. In Mymensingh, it reached its peak F1-score of \textbf{82.611\%} at \textbf{epoch 20}, while in Sylhet, it achieved \textbf{82.315\%} at \textbf{epoch 20}, demonstrating its ability to effectively handle regional dialects. In Chittagong, the model outperformed the others, with a peak F1-score of \textbf{76.377\%} at \textbf{epoch 20}, indicating its proficiency in recognizing named entities in this dialect. However, in Barishal and Noakhali, the performance was more modest, with Barishal achieving an F1-score of \textbf{77.863\%} at \textbf{epoch 15} and Noakhali reaching \textbf{81.553\%} at \textbf{epoch 20}. These results highlight the model's strength in multilingual contexts while also suggesting that further fine-tuning could improve it's performance in specific regional dialects.

\begin{figure}[htbp]
    \centering
    % First Image - Chittagong
    \begin{subfigure}[b]{0.48\textwidth}
        \centering
        \includegraphics[width=\textwidth, keepaspectratio]{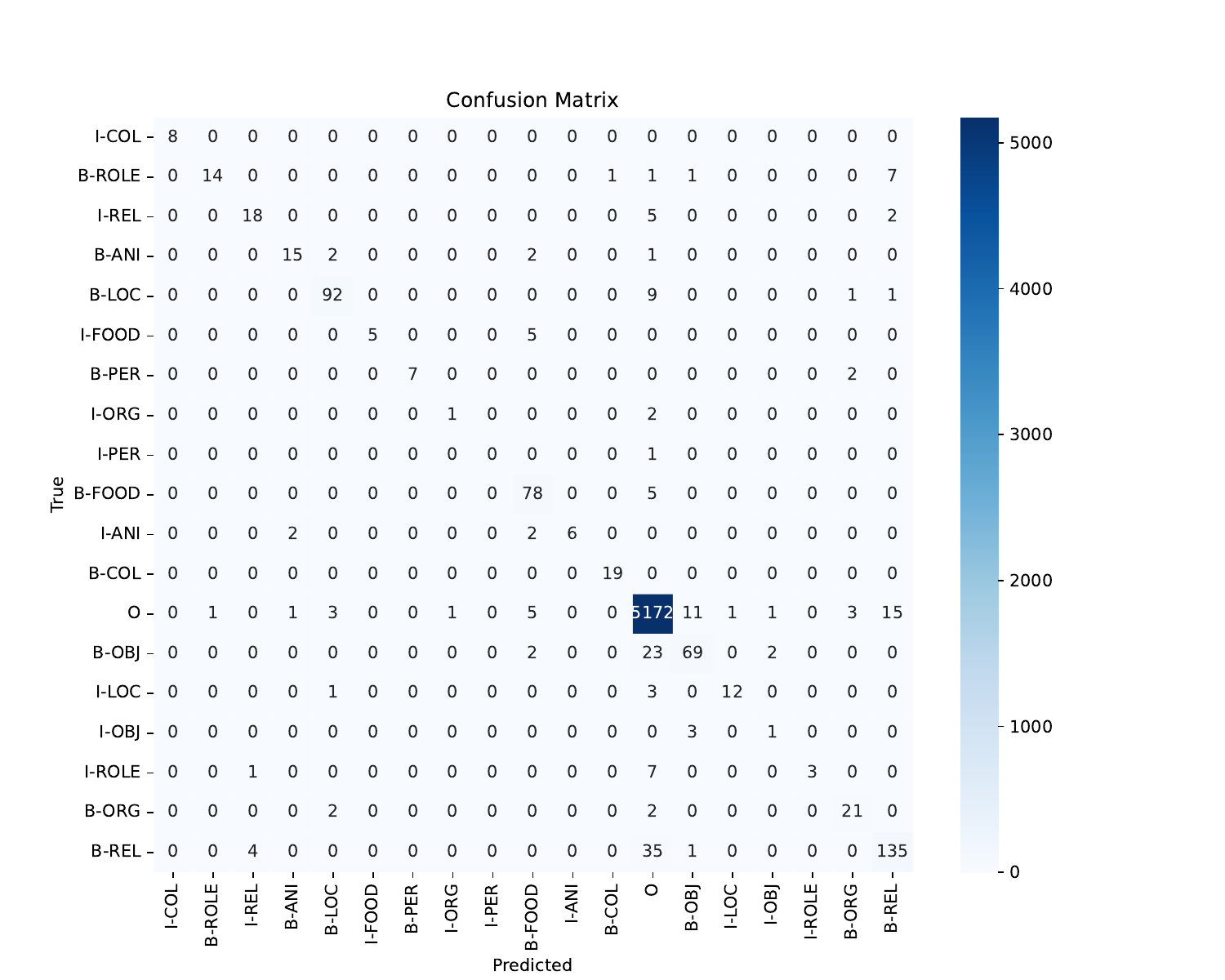} % Replace with your image file
        \caption{Barishal}
    \end{subfigure}
    \hspace{0.02\textwidth} % Adjust the spacing between the subfigures if needed
    % Second Image - Sylhet
    \begin{subfigure}[b]{0.48\textwidth}
        \centering
        \includegraphics[width=\textwidth, keepaspectratio]{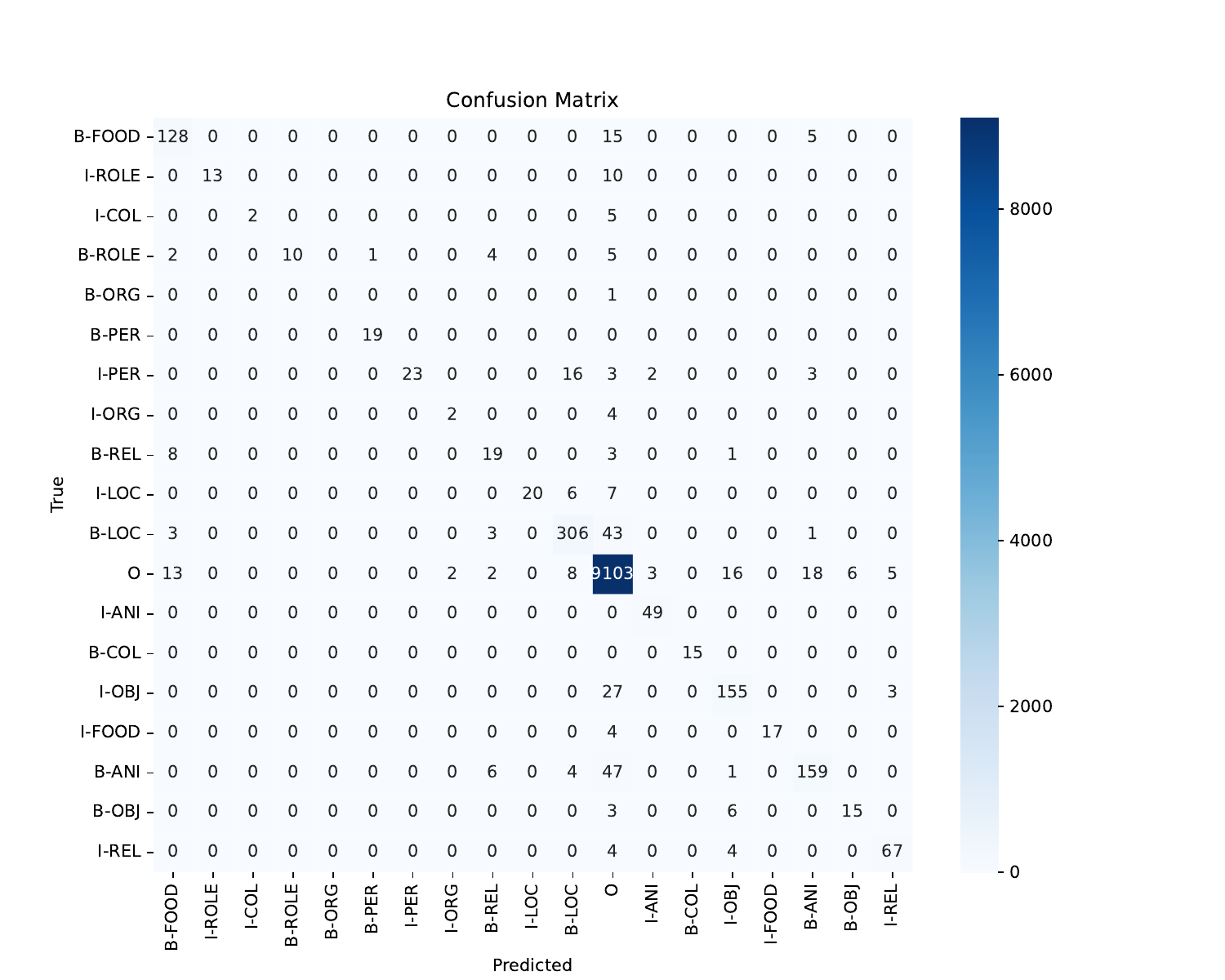} % Replace with your image file
        \caption{Mymensingh}
    \end{subfigure}
    
    % Third Image - Barishal
    \begin{subfigure}[b]{0.48\textwidth}
        \centering
        \includegraphics[width=\textwidth, keepaspectratio]{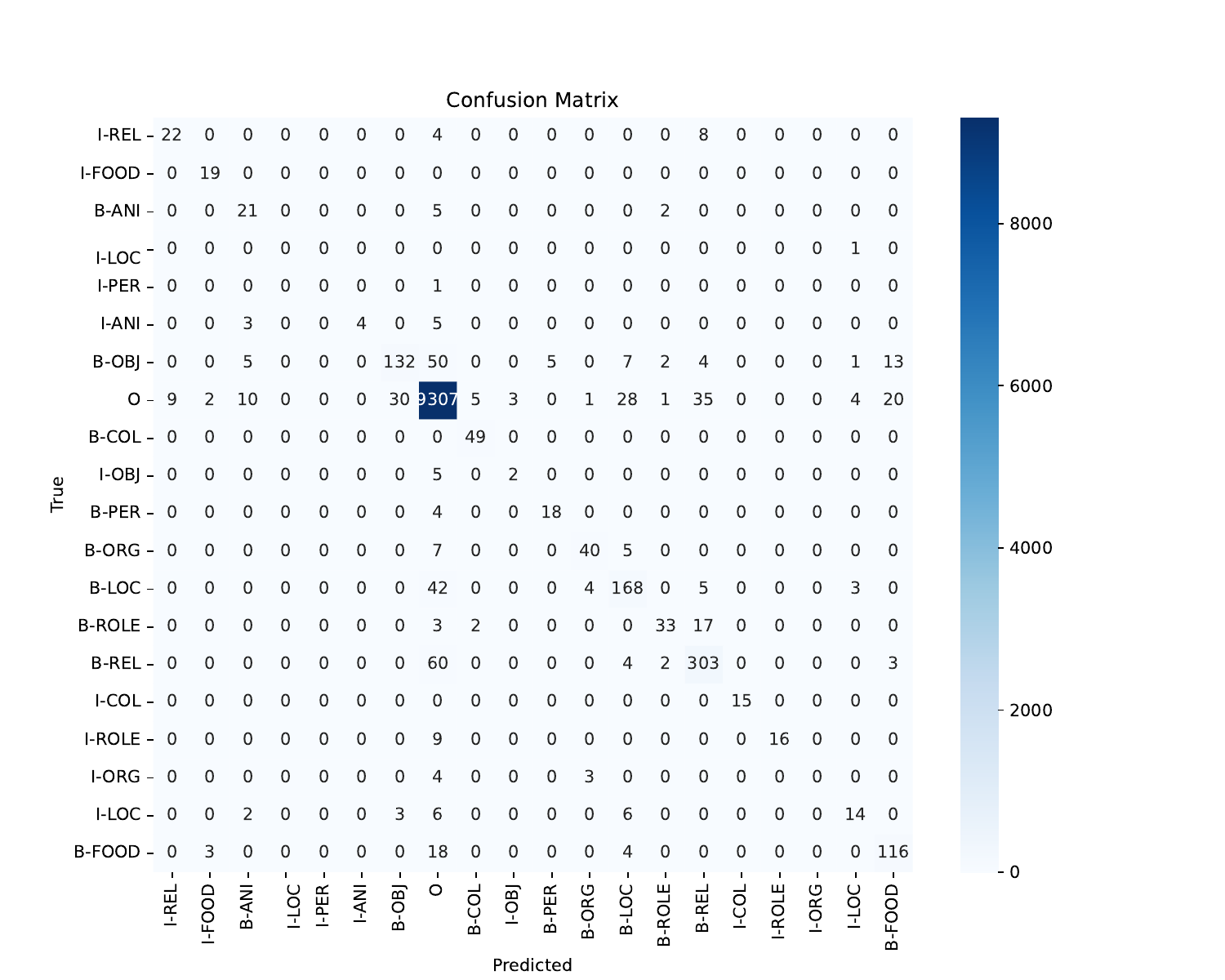} % Replace with your image file
        \caption{Chittagong}
    \end{subfigure}
    \hspace{0.02\textwidth} % Adjust the spacing between the subfigures if needed
    % Fourth and Fifth Image - Noakhali and Mymensingh (Side by Side)
    \begin{subfigure}[b]{0.48\textwidth}
        \centering
        \includegraphics[width=\textwidth, keepaspectratio]{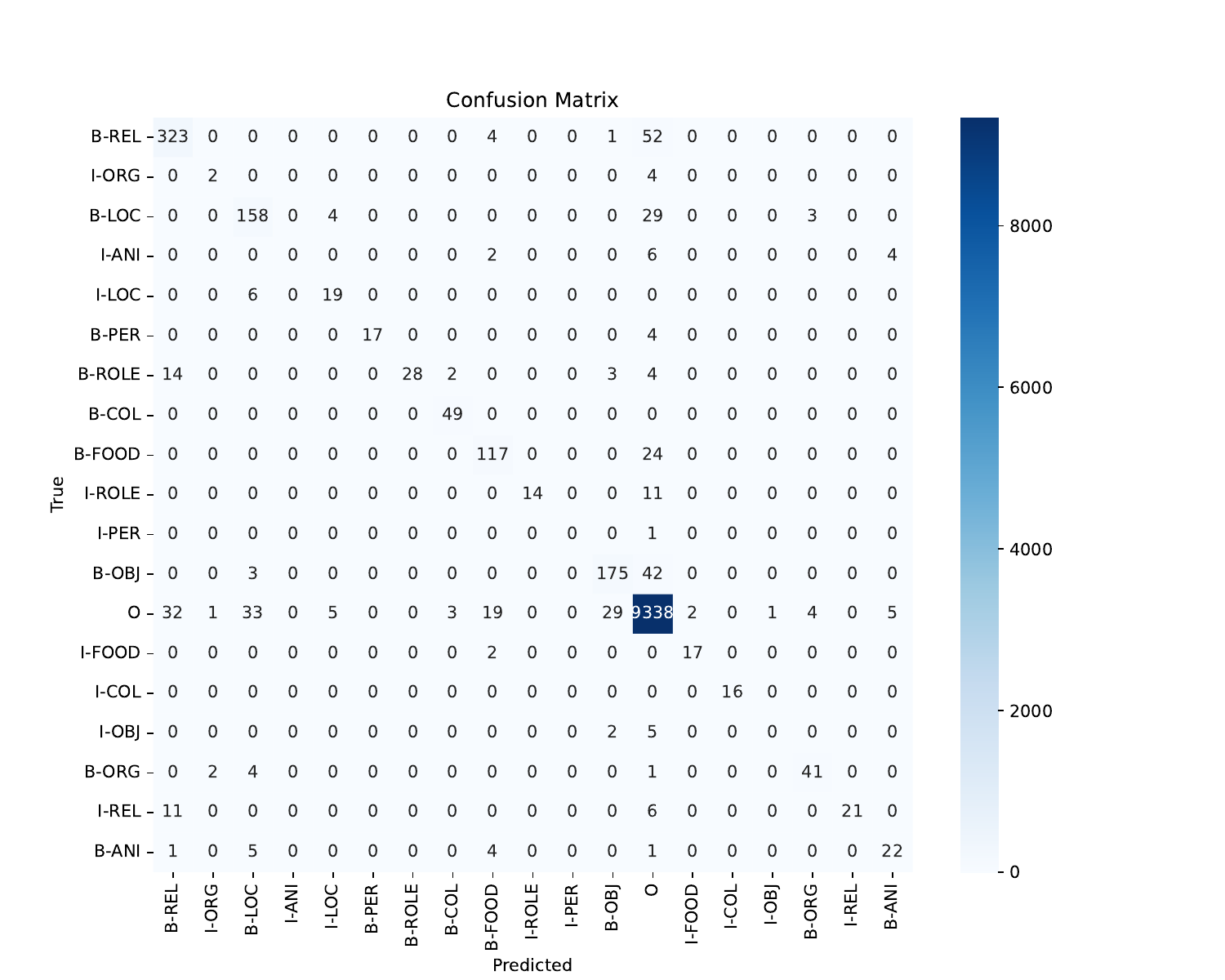} % Replace with your image file
        \caption{Noakhali}
    \end{subfigure}

    \begin{subfigure}[b]{0.48\textwidth}
        \centering
        \includegraphics[width=\textwidth, keepaspectratio]{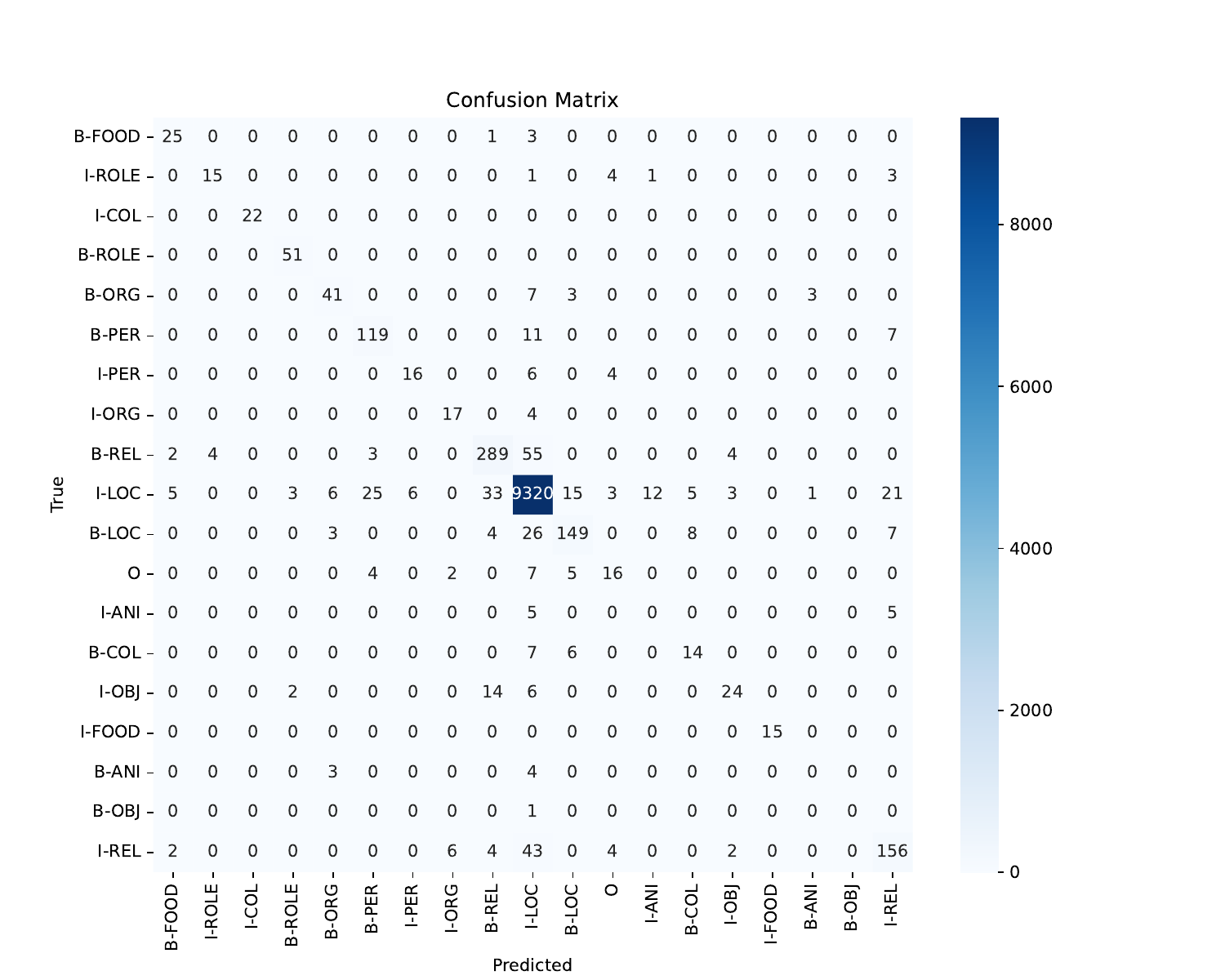} % Replace with your image file
        \caption{Sylhet}
    \end{subfigure}
    
    \caption{Confusion Matrices for the Best Performing Model Across Five Bangla Regional Dialects}
    \label{fig:confusion}
\end{figure}
%Old
% \begin{figure}[h]
%     \centering
%     % First Image - Chittagong
%     \begin{minipage}{0.42\textwidth}
%         \centering
%         \includegraphics[width=\linewidth]{Confusion/con ba.pdf} % Replace with actual image file path
%         \subcaption{Barishal}
%     \end{minipage}
%     \hfill
%     % Second Image - Sylhet
%     \begin{minipage}{0.42\textwidth}
%         \centering
%         \includegraphics[width=\linewidth]{Confusion/con_my.pdf} % Replace with actual image file path
%         \subcaption{Mymensingh}
%     \end{minipage}
%     \hfill
%     % Third Image - Barishal
%     \begin{minipage}{0.42\textwidth}
%         \centering
%         \includegraphics[width=\linewidth]{Confusion/confusion_matrix_Chittagong.pdf} % Replace with actual image file path
%         \subcaption{Chittagong}
%     \end{minipage}
%     \hfill % Space between the two rows
%     % Fourth and Fifth Image - Noakhali and Mymensingh (Side by Side)
%     \begin{minipage}{0.42\textwidth}
%         \centering
%         \includegraphics[width=\linewidth]{Confusion/con_noa.pdf} % Replace with actual image file path
%         \subcaption{Noakhali}
%     \end{minipage}
%     \hfill
%     \begin{minipage}{0.42\textwidth}
%         \centering
%         \includegraphics[width=\linewidth]{Confusion/confusion_sy.pdf} % Replace with actual image file path
%         \subcaption{Sylhet}
%     \end{minipage}
%     \caption{Confusion Matrices for the Best Performing Model Across Five Bangla Regional Dialects}
%     \label{fig:confusion}
% \end{figure}

In the Figure~\ref{fig:confusion}, the confusion matrices for the Barishal, Chittagong, Mymensingh, Noakhali, and Sylhet regions provide valuable insights into the performance of the NER models. In the Barishal region, the model performs well in recognizing Location (LOC) and Food (FOOD) entities, with relatively few misclassifications. However, categories like Role (ROLE) and Organization (ORG) show higher rates of false positives, indicating that the model sometimes confuses these entities with others. The Chittagong region's confusion matrix reveals similar patterns, with good recognition of Person (PER) and Location (LOC) entities but a noticeable tendency to misclassify Role (ROLE) and Organization (ORG) entities. Mymensingh demonstrates strong precision across all entities, particularly for Location (LOC) and Person (PER), but also reveals a few misclassifications in other categories. Noakhali and Sylhet regions exhibit somewhat lower performance, especially for Organization (ORG) and Role (ROLE) categories, where the false positives are more prominent. Overall, the confusion matrices highlight that while the model excels in recognizing some entities across all regions, there are specific areas—particularly in Role (ROLE) and Organization (ORG)—that require further refinement for improved accuracy.

In conclusion, the performance analysis of the three models—Bangla BERT, Bangla BERT Base, and BERT Base Multilingual Cased—demonstrates that while BERT Base Multilingual Cased generally performed the best across most regions, Bangla BERT also showed strong results in certain areas. Mymensingh achieved its highest F1-score of \textbf{82.611\%}with BERT Base Multilingual Cased at \textbf{epoch 20}, while Sylhet performed best with BERT Base Multilingual Cased, reaching an F1-score of \textbf{82.315\%}. Barishal saw the highest performance with Bangla BERT, with an F1-score of \textbf{81.481\%} at \textbf{epoch 20}. Chittagong and Noakhali, however, exhibited more modest improvements, with Chittagong achieving the highest F1-score of \textbf{75.307\%} with Bangla BERT at \textbf{epoch 20}, and Noakhali performing best with BERT Base Multilingual Cased, achieving an F1-score of \textbf{81.553\%} at \textbf{epoch 20}. These findings suggest that while BERT Base Multilingual Cased excels in handling multilingual contexts and regional variations, Bangla BERT is more effective in certain regions, particularly in Mymensingh and Barishal. The confusion matrices further highlight the model's strengths and areas for improvement, especially in Role (ROLE) and Organization (ORG) categories, where false positives were more frequent across regions. Overall, the results indicate that further fine-tuning and region-specific adjustments are necessary to enhance the models' performance, especially in more complex dialects.

\section{Conclusion and Future work}
\label{Conclusion_and_Future_work}

In this study, we introduced ANCHOLIK-NER, the first benchmark dataset for Named Entity Recognition (NER) in Bangla regional dialects, specifically focusing on Sylhet, Chittagong, Barishal, Noakhali, and Mymensingh. The dataset was developed to bridge the gap in NER resources for Bangla dialects, which have been underrepresented in computational linguistics. Our results demonstrate that the Bangla BERT model, fine-tuned specifically for Bangla, outperforms the other models, including Bangla BERT Base and BERT Base Multilingual Cased, in recognizing named entities across all regions, particularly in Mymensingh and Barishal. Despite the strong overall performance, regions like Chittagong and Noakhali presented challenges, with relatively lower precision and recall scores, suggesting the need for further fine-tuning or additional data for these regions. The proposed ANCHOLIK-NER dataset provides a valuable resource for training and evaluating NER models tailored to Bangla regional dialects, and it can contribute to the development of more inclusive and dialect-aware NLP systems. The findings from this study underline the importance of using region-specific models to improve NER performance and highlight the potential for further research in dialectal variations within the Bangla language.

While ANCHOLIK-NER fills a significant gap in NER resources for Bangla regional dialects, several challenges remain. First, the dataset currently covers only five regions, leaving many other dialects and sub-dialects underrepresented. Further expansion of the dataset to include more diverse regions and their specific linguistic features would enhance the generalizability of the models. Additionally, though the current study uses existing transformer models, future research could explore the potential of more advanced techniques, such as incorporating hybrid models or leveraging unsupervised learning approaches to capture dialectal variations more effectively. Moreover, despite the strong results for most regions, Chittagong and Noakhali showed relatively lower performance, indicating that more refined techniques, such as domain adaptation or regional-specific data augmentation, are necessary to address these specific dialects' challenges.

Future work will focus on enhancing the performance of the models in regions like Chittagong and Noakhali, where the models showed lower accuracy. This can be achieved through additional training data, further fine-tuning, or the incorporation of region-specific linguistic features. Additionally, exploring more advanced pre-training techniques or leveraging other language models for Bangla, such as multilingual transformers with fine-tuning on dialect-specific corpora, could improve the overall model performance. Furthermore, expanding the dataset to include more dialects and regional variations could help in building a more robust NER system that performs well across all Bangla dialects.

% \section{Acknowledgments}

% This research did not receive any specific grant from funding agencies in the public, commercial, or not-for-profit sectors.

\section{Author Contributions}
\textbf{Bidyarthi Paul:} Methodology, Data curation, Writing – original draft, Visualization, Validation

\textbf{Faika Fairuj Preotee:} Data curation, Investigation, Writing – original draft

\textbf{Shuvashis Sarker:} Methodology, Validation,  Software, Conceptualization

\textbf{Shamim Rahim Refat:} Resources, Data curation, Project administration

\textbf{Shifat Islam:} Supervision, Project administration

\textbf{Tashreef Muhammad:} Supervision, Writing – review \& editing

\textbf{Mohammad Ashraful Hoque:} Project administration

\textbf{Shahriar Manzoor:} Supervision.

\bibliographystyle{unsrt}  
\bibliography{references}  %%% Remove comment to use the external .bib file (using bibtex).

\end{document}